\documentclass[10pt, journal,twoside]{IEEEtran}
\usepackage{amsmath,amsfonts}
\usepackage{algorithmic}
\usepackage{array}
\usepackage[caption=false,font=normalsize,labelfont=sf,textfont=sf]{subfig}
\usepackage{textcomp}
\usepackage{stfloats}
\usepackage{url}
\usepackage{verbatim}
\usepackage{graphicx}
\usepackage{ulem}
\usepackage{cite}
\usepackage{booktabs}
\usepackage[T1]{fontenc}
\usepackage[utf8]{inputenc} 
\usepackage[square,sort&compress,numbers]{natbib}
\def\BibTeX{{\rm B\kern-.05em{\sc i\kern-.025em b}\kern-.08em
    T\kern-.1667em\lower.7ex\hbox{E}\kern-.125emX}}
\usepackage{balance}
\usepackage[table,xcdraw,dvipsnames]{xcolor} 
\usepackage{color}
\definecolor{citecolor}{RGB}{66,168,235}
\definecolor{linkcolor}{RGB}{255,0,0}
\usepackage{hyperref}
\hypersetup{colorlinks=true,citecolor=citecolor,linkcolor=linkcolor,urlcolor=Thistle}

\begin{document}
\title{SPEX: A Vision-Language Model for Land Cover Extraction on Spectral Remote Sensing Images}
\author{Dongchen Si, Di Wang, Erzhong Gao, Xiaolei Qin, Liu Zhao, Jing Zhang,~\IEEEmembership{Senior Member,~IEEE}, Minqiang Xu, \\Jianbo Zhan, Jianshe Wang, Lin Liu, Bo Du,~\IEEEmembership{Senior Member,~IEEE} and Liangpei Zhang,~\IEEEmembership{Fellow,~IEEE}

\thanks{Dongchen Si and Di Wang contributed equally to this paper. \textit{(Corresponding author: Minqiang Xu and Jianbo Zhan)}}

\thanks{Dongchen Si is with the College of Computer Science and Technology, Xinjiang University, Urumqi 830046, China, the iFlytek Co., Ltd., Hefei 230088, China, and also with the National Engineering Research Center of Speech and Language Information Processing, Hefei 230088, China (e-mail: dochsi@outlook.com).}

\thanks{Di Wang, Jing Zhang, and Bo Du are with the School of Computer Science, Wuhan University, Wuhan 430072, China, and also with the Zhongguancun Academy, Beijing 100094, China. In addition, Di Wang and Bo Du are with the National Engineering Research Center for Multimedia Software, Hubei Key Laboratory of Multimedia and Network Communication Engineering, Wuhan University, Wuhan 430072, China (e-mail: wd74108520@gmail.com; jingzhang.cv@gmail.com; dubo@whu.edu.cn).}

\thanks{Erzhong Gao, Jianbo Zhan, Liu Zhao, Jianshe Wang, Lin Liu, and Minqiang Xu are with the iFlytek Co., Ltd., Hefei, Anhui 230088, China and also with the National Engineering Research Center of Speech and Language Information Processing, Hefei, Anhui 230088, China (e-mail: ezgao@iflytek.com; jbzhan@iflytek.com; zaynmalik00612@gmail.com; sir816@163.com; linliu@iflytek.com; mqxu7@iflytek.com).}

\thanks{Xiaolei Qin and Liangpei Zhang are with the State Key Laboratory of Information Engineering in Surveying, Mapping and Remote Sensing, Wuhan University, Wuhan, Hubei 430079, China (e-mail: qinxlei@whu.edu.cn; zlp62@whu.edu.cn).}
}
\markboth{Journal of \LaTeX\ Class Files,~Vol.~18, No.~9, September~2020}{SPEX for Land Cover Extraction}

\maketitle
\begin{abstract}

Spectral information has long been recognized as a critical cue in remote sensing observations. Although numerous vision-language models have been developed for pixel-level interpretation, spectral information remains underutilized, resulting in suboptimal performance, particularly in multispectral scenarios. To address this limitation, we construct a vision-language instruction-following dataset named SPIE, which encodes spectral priors of land-cover objects into textual attributes recognizable by large language models (LLMs), based on classical spectral index computations. Leveraging this dataset, we propose SPEX, a multimodal LLM designed for instruction-driven land cover extraction. To this end, we introduce several carefully designed components and training strategies, including multiscale feature aggregation, token context condensation, and multispectral visual pre-training, to achieve precise and flexible pixel-level interpretation. To the best of our knowledge, SPEX is the first multimodal vision-language model dedicated to land cover extraction in spectral remote sensing imagery. Extensive experiments on five public multispectral datasets demonstrate that SPEX consistently outperforms existing state-of-the-art methods in extracting typical land cover categories such as vegetation, buildings, and water bodies. Moreover, SPEX is capable of generating textual explanations for its predictions, thereby enhancing interpretability and user-friendliness. Code will be released at: \href{https://github.com/MiliLab/SPEX}{https://github.com/MiliLab/SPEX}.

\end{abstract}

\IEEEpeerreviewmaketitle
\section{Introduction}

\IEEEPARstart{R}emote sensing technology enables the collection, processing, and imaging of electromagnetic waves reflected or emitted by targets from a distance. It utilizes sensors mounted on aircraft or satellites to record electromagnetic radiation from the Earth's surface. Beyond capturing visual features, remote sensing images contain rich spectral information ranging from visible light to shortwave infrared bands, which facilitates accurate identification of various land objects. Leveraging this advantage, remote sensing imagery has been widely applied in diverse fields such as environmental monitoring, resource management, and disaster assessment. Among these applications, land cover extraction, i.e., pixel-level classification of specific land cover types, is a particularly crucial and foundational task, especially when working with multispectral images that span extensive geographic areas.

\begin{figure}[t]
  \centering
    \includegraphics[width=\linewidth]{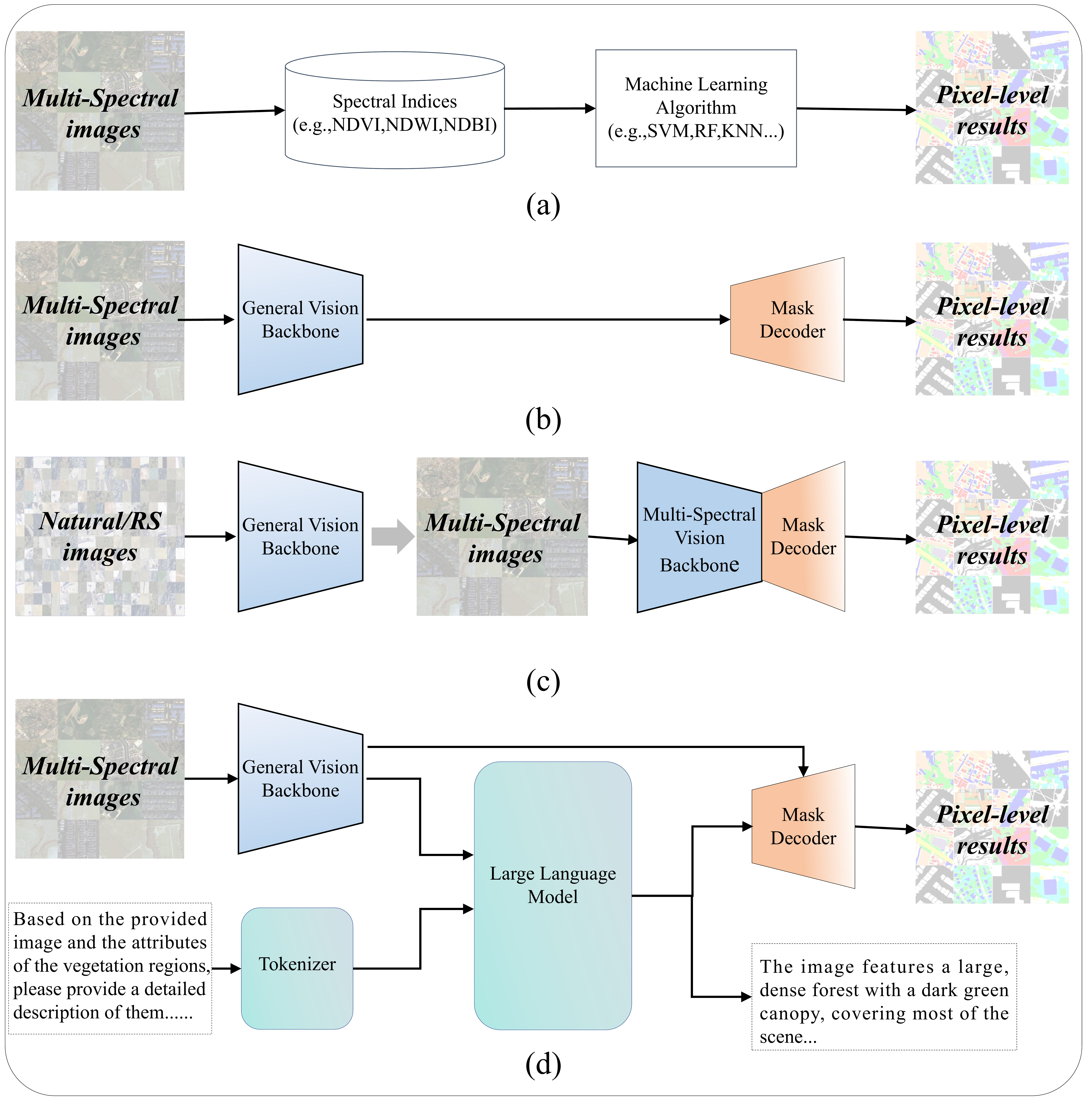}
    \caption{Comparison of various methods for land cover extraction. (a) combines traditional spectral index techniques with machine learning methods; (b) directly trains segmentation networks on multispectral images; (c) first pre-trains visual models on large-scale images, which are then used as backbones for segmentation networks; (d) further extends this approach by introducing a language model, enabling flexible and interactive interpretations guided by textual instructions.}\label{pattern_comparison}
\end{figure}%

Existing land feature extraction methods can be broadly categorized into two groups: traditional approaches and deep learning-based methods. Traditional techniques include single-band thresholding \citep{du2012estimating}, spectral relationship analysis \citep{sawaya2003extending}, and index-based methods \citep{zeng2019extracting}. These approaches primarily rely on the spectral characteristics of various bands in remote sensing imagery and achieve feature identification and extraction through manually defined rules, empirical thresholds, or automated thresholding algorithms \citep{liu2022composition}. However, such manual feature engineering methods often lack robustness and exhibit unstable performance in complex environments \citep{li2023object}. As a result, they are increasingly being supplanted by deep learning approaches \citep{xu2024multi}, which significantly enhance task performance by automatically learning inherent semantic features from imagery. In particular, inspired by the success of vision foundation models on natural scene understanding \citep{nguyen2024image}, numerous vision networks pre-trained on remote sensing imagery have emerged as a hot research topic \citep{wang2025hypersigma, wang2024mtp}. These models have demonstrated impressive performance across a range of downstream remote sensing tasks, including land cover extraction \citep{hong2024spectralgpt}.

Nevertheless, current remote sensing visual models mainly produce static interpretation results and lack interactivity. They cannot refine outputs based on user feedback or flexibly adapt to diverse interpretation needs, such as user-defined regions or specific object categories. Moreover, changing the target category typically requires retraining the model, leading to low efficiency. In addition, the results are often only interpretable by domain experts, which severely limits their accessibility and practical usability.

The emergence of large language models (LLMs) \citep{guo2025deepseek} offers a promising solution to the aforementioned challenges. With strong natural language understanding, multi-turn interaction, and advanced reasoning capabilities, LLMs exhibit a high degree of interactive intelligence, enabling more accurate and personalized user assistance. Consequently, LLMs substantially expand the potential of applications such as intelligent question answering and decision support.

In this context, integrating remote sensing vision models with large language models to develop remote sensing vision-language models (RSVLMs) has emerged as a key trend \citep{shabbir2025geopixel,zhan2025skyeyegpt,hu2025rsgpt}. These models combine visual perception with natural language understanding, enabling not only the processing of remote sensing imagery but also supporting tasks such as description, question answering, reasoning, and decision-making (see Figure \ref{pattern_comparison}). This integration overcomes the limitations of previous remote sensing visual models that “perception but do not understand”, and equips remote sensing systems with enhanced cognitive capabilities and greater potential for human-computer interaction. For example, Wu et al. \citep{wu2025fsvlm} propose a dual-branch framework that integrates LLaVA \citep{li2023llava} and SAM \citep{kirillov2023segment} for language-guided farmland segmentation.

Despite the significant progress achieved by recent RSVLMs, several critical challenges remain in effectively processing multispectral imagery. First, the spatial representation of RSVLMs is inherently constrained by sensor resolution. For typical multispectral remote sensing data, such as Landsat-8 (30 m) and Gaofen-2 (10 m), this often leads to blurred textures and the loss of fine-grained structures, making accurate land object recognition particularly challenging. Second, most existing methods still follow a three-channel (RGB) modeling paradigm and lack explicit representations of spectral responses beyond RGB as well as reflectance relationships across different bands, which limits their ability to fully exploit the rich spectral information available in multispectral imagery. Motivated by these observations, we propose SPEX, short for SPectral instruction EXtraction, a novel RSVLM that integrates LLMs with cross-band spectral information to enable instruction-based land-cover extraction from multispectral remote sensing images. Specifically, SPEX combines a vision encoder, an LLM, and a lightweight decoder into an end-to-end multimodal architecture capable of pixel-level interpretation. To alleviate the performance degradation caused by resolution constraints, we adopt a hierarchical vision encoder and introduce a multi-scale feature aggregation strategy, which aims to preserve fine-grained spatial structures and more effectively exploit visual information for land-cover extraction. In parallel, to address the insufficient utilization of intrinsic spectral information in existing RSVLMs, we construct a novel multimodal dataset termed Spectral Prompt Instruction Extraction (SPIE). Unlike generic vision–language datasets, each SPIE sample pairs multispectral imagery with textual instructions enriched by object attributes derived from spectral computations across different bands, thereby facilitating more interpretable and robust semantic understanding of land-cover categories in multispectral imagery. Extensive experiments on multiple public datasets demonstrate that SPEX not only achieves state-of-the-art performance compared with existing RSVLMs, but also provides improved interpretability for land-cover extraction results.

The main contributions of this paper can be summarized as follows:

\begin{itemize}

    \item[(1)] We propose SPEX, the first vision-language model tailored for instruction-based pixel-level land cover extraction from spectral remote sensing imagery. By leveraging spectral information and the capabilities of large language models, SPEX enables high-precision land cover extraction through spectral prompt-based instructions.

    \item[(2)] We construct a spectral prompt instruction extraction dataset that leverages spectral priors and auxiliary instructions to transform multispectral imagery into vision–language pairs enriched with textual object attributes and descriptions, thereby strengthening SPEX’s ability to recognize remote sensing objects.

     \item[(3)] We conduct extensive experiments on five remote sensing datasets covering typical land cover categories such as vegetation, water bodies, and buildings. The results demonstrate that SPEX achieves state-of-the-art performance compared to existing methods. In addition, qualitative analyses highlight the explainability of SPEX for multispectral object extraction.
    
\end{itemize}
\section{Related Work}

\subsection{Land Cover Extraction}

Land cover extraction is a fundamental task of remote sensing interpretation.
Early research mainly focused on spectral index methods \cite{bouhennache2019new,gu2018building}. These methods target the spectral reflectance characteristics of land objects by constructing diverse spectral indices through various band combinations. \cite{zhang2021fusion} and \cite{zeng2019extracting} further integrated spectral index methods with traditional machine learning classifiers, improving the accuracy and adaptability of land cover extraction. Subsequently, deep learning methods have emerged and gradually supplanted traditional approaches, becoming the dominant paradigm in current research \cite{xiong2024neural, cong2022satmae, szwarcman2024prithvi, hong2024spectralgpt}. These methods not only enable automatic extraction of spatial and spectral features but also facilitate adaptive learning of contextual semantic information, which is essential for accurate understanding of remote sensing imagery. Nevertheless, these approaches usually exhibit limited generalization and are difficult to adapt to diverse scenes through natural language instructions. This limitation arises from their primary reliance on visual-only models trained under task-specific supervision with fixed extraction objectives, which restricts their ability to dynamically adjust extraction behavior according to user intent. In contrast, this paper proposes SPEX, a novel framework that introduces LLMs to enable text-guided, pixel-level extraction of land cover targets from multispectral remote sensing images, thereby improving the flexibility and practicality of land cover interpretation.

\subsection{Gereral Vision-Language Model}

Vision-Language Models (VLMs) have significantly advanced human-computer interaction and activated the image understanding capabilities of LLMs, enabling visual information to be more effectively interpreted in textual form. For example, models like Flamingo \cite{alayrac2022flamingo}, LLaVA \cite{li2023llava}, Instruct-BLIP \cite{Dai2023InstructBLIPTG}, and MiniGPT-4 \cite{zhu2023minigpt} have achieved remarkable results in visual question answering and image-text multimodal dialogue tasks, showcasing the potential of VLMs in cross-modal understanding. Nevertheless, these models typically struggle to perform pixel-level interpretation. Recently, LISA \cite{lai2024lisa} established a new paradigm by introducing a [SEG] token to bridge LLMs with segmentation decoders such as SAM, enabling language-guided mask prediction. Subsequent studies further refined this paradigm: GSVA \cite{xia2024gsva} introduced shared-weight [SEG] and [REJ] tokens to address multi-target and empty-target scenarios; GLaMM \cite{rasheed2024glamm} enabled pixel-grounded conversational capabilities through holistic segmentation, while OMG-LLaVA \cite{zhang2024omg} integrated a general segmentation backbone with LLMs to support pixel-level reasoning. Despite their success, these general-purpose VLMs are not explicitly adapted to remote sensing data characteristics, such as top-down imaging conditions and large-scale scene structures, which are crucial for accurate land object identification.

\subsection{Remote Sensing Vision-Language Model} 

Most existing RSVLMs are developed by fine-tuning general VLMs on constructed remote sensing image-text datasets. For example, RSGPT \cite{hu2025rsgpt} builds upon InstructBLIP \cite{Dai2023InstructBLIPTG} and introduces manually annotated instruction-tuning texts based on remote sensing images. EarthGPT \cite{zhang2024earthgpt} constructs a large-scale multimodal instruction-following dataset for remote sensing by converting the labels of public datasets into question-answer pairs. GeoChat \cite{kuckreja2024geochat}, built upon LLaVA-1.5 \cite{liu2023visual}, implements an automated annotation pipeline by converting object attributes into instruction-style texts using GPT, significantly reducing the cost of dataset construction. Other models also have their own specific focuses. For example, SkyEyeGPT \cite{zhan2025skyeyegpt} emphasizes multi-turn dialogue, SkySenseGPT \cite{luo2024skysensegpt} focuses on the relationships among scene objects, LHRS-Bot \cite{muhtar2024lhrs} adopts curriculum learning for progressive training, and H²RSVLM \cite{pang2024h2rsvlm} enhances model honesty. Nevertheless, these models are primarily designed for image-level semantic understanding and dialogue, and do not explicitly support pixel-level interpretation, which limits their applicability to land-cover extraction and object segmentation tasks. To this end, several recent works incorporate segmentation-aware extensions. RSUniVLM \cite{liu2024rsunivlm} introduces a granularity-oriented mixture-of-experts mechanism, GeoPixel \cite{shabbir2025geopixel} generates extensive textual annotations using segmentation masks and spatial priors tailored for remote sensing data, and GeoPix \cite{ou2025geopix} employs a class-wise learnable memory in the decoder for adaptive mask learning. For land cover extraction, FSVLM \cite{wu2025fsvlm} and FarmSeg\_VLM \cite{wu2025farmseg_vlm} adopt the "embedding-as-mask" paradigm to extract farmland regions. However, most of these approaches are still primarily designed under an RGB-centric assumption and lack explicit mechanisms to model inter-channel spectral relationships, making them difficult to directly extend to multispectral imagery. In contrast, SPEX is specifically designed for multispectral remote sensing images and explicitly models cross-band spectral relationships, while integrating language-driven visual perception and reasoning to enable accurate and interpretable land-cover extraction.

\section{Methodology}

In this section, we present the details of the proposed method. We begin by introducing the construction process of the training dataset SPIE. Then, we systematically describe the overall workflow, internal components, and training strategy of SPEX.

\subsection{Spectral Prompt Instruction Extraction Dataset}

Current RSVLMs are largely driven by the quality of constructed image-text datasets. To enable instruction-based land cover extraction on multispectral imagery, we curate a multimodal instruction-following dataset that explicitly leverages the spectral characteristics of remote sensing objects. As the spectral knowledge is embedded into the instruction via textual prompts, we refer to this dataset as the Spectral Prompt Instruction Extraction (SPIE).

Each sample in SPIE consists of four components: the original image, a binary extraction map, a task-specific instruction, and the corresponding response. The remainder of this section details the construction process of SPIE, including image collection, instruction and response generation, and spectral knowledge integration.

\begin{table*}[t]
\centering
\caption{Detailed information of different sub-datasets in SPIE. NIR* can be inferred from NDVI images in Globe230k.}
\begin{tabular}{l c l c c c c c}
\toprule
\textbf{Dataset} & \textbf{Resolution}  & \textbf{Bands} & \textbf{Training Samples} & \textbf{Testing Samples} & \textbf{Vegetation} & \textbf{Building} & \textbf{Water} \\
\midrule
SegMunich \cite{hong2024spectralgpt} & 128$\times$128  & R, G, B, NIR & 39,402 & 9,841 & $\checkmark$ & $\times$ & $\times$ \\
Chesapeake \cite{robinson2019large} & 512$\times$512  & R, G, B, NIR & 154,037 & 5,525 & $\checkmark$ & $\times$ & $\times$ \\
Globe230k \cite{shi2023globe230k} & 512$\times$512  & R, G, B, NIR* & 137,064 & 15,470 & $\checkmark$ & $\times$ & $\times$ \\
SpaceNet-V2 \cite{van2018spacenet} & 650$\times$650  & R, G, B, NIR, SWIR & 8,474 & 2,119 & $\times$ & $\checkmark$ & $\times$ \\
GID-15 \cite{tong2020land} & 512$\times$512  & R, G, B, NIR & 25,199 & 6,301 & $\times$ & $\times$ & $\checkmark$ \\
\bottomrule
\end{tabular}%
\small
\\
\vspace{0.5em}
\textit{R}: Red, \textit{G}: Green, \textit{B}: Blue, \textit{NIR}: Near Infrared, \textit{SWIR}: Short Wave Infrared \label{table1}
\end{table*}

\subsubsection{Image Collection}

\noindent\textbf{Data Source} To enhance the model’s representation capacity and generalization ability, we jointly utilize images from five multispectral semantic segmentation datasets: SegMunich \cite{hong2024spectralgpt}, Chesapeake \cite{robinson2019large}, Globe230K \cite{shi2023globe230k}, GID \cite{tong2020land}, and SpaceNet-V2 \cite{van2018spacenet}. Since the original image sizes in Chesapeake and GID-15 are relatively large, the images are cropped into 512 $\times$ 512 patches to ensure model training. Nevertheless, the resulting samples still adhere to the original train-test split. For the other datasets, images are directly used as provided, with the official dataset splits preserved.

\noindent\textbf{Binary Extraction Map} To obtain high-quality object extraction mask labels for model training and evaluation, we utilize the segmentation maps provided in the five aforementioned datasets, where each pixel is annotated with a specific semantic class label, offering precise geospatial information.

In this study, we focus on three primary land cover types: vegetation, water, and buildings, due to their importance in practical applications. Specifically, SegMunich, Chesapeake, and Globe230K are used for vegetation extraction, while GID and SpaceNet-V2 provide extensive annotations for water bodies and buildings, respectively. For these purposes, we derive binary masks for the corresponding categories based on the original segmentation labels.

Since category definitions vary across datasets (e.g., forests and grasslands are treated as separate classes), we unify related subclasses into a single vegetation category, enabling a binary classification of vegetation vs. non-vegetation. The same merging strategy is applied to water and building categories to ensure consistency across datasets. Guided by this principle, in the Globe230K dataset, the classes “forest,” “grassland,” and “shrubland” are merged into the vegetation category. Similarly, in the GID dataset, “river,” “lake,” and “pond” are unified under the water category.

Through this meticulous selection and integration process, we construct a high-quality binary extraction mask dataset targeting specific land cover types, which serves as a visual foundation for subsequent model training. Detailed information for each subset is provided in Table~\ref{table1}.

\subsubsection{Instruction and Response Generation}

After constructing the visual component of SPIE, we proceed to generate the corresponding textual content, which consists of two parts: the instruction and the response.

In this study, the instructions are designed to contain two elements: a query template and a spectral prompt. The spectral prompts are expressed in natural language and embed the spectral characteristics of the multispectral image. To ensure consistency, we adopt a fixed instruction template:

\begin{quote}
\textit{Based on the provided image and the attributes of the [Land Cover Category] regions, please provide a detailed description of them: [Spectral Prompt]}
\end{quote}

Here, [Land Cover Category] refers to one of the target types: vegetation, building, or water, while the [Spectral Prompt] conveys key spectral attributes specific to the selected category. The construction of these spectral prompts will be detailed in the next section.

Given the image and instruction, generating the response is relatively straightforward. In this study, the response is expected to provide a comprehensive description of the image content, with a particular focus on the specified land cover category. Intuitively, the response can be obtained directly by applying existing advanced MLLMs to the image–instruction pair. However, this approach faces two key challenges. First, the spectral attributes embedded in the prompt are often difficult for general-purpose MLLMs to interpret accurately, which may degrade the quality of the generated descriptions. Second, using MLLM-generated responses as training targets based on the same inputs, can introduce an inductive bias into SPEX, potentially reducing its generalization capability.

To address this challenge, we propose a novel strategy that leverages auxiliary instructions to guide response generation. Specifically, we generate descriptive responses using additional instructions that are not involved in the training of SPEX. These auxiliary instructions are carefully crafted to elicit well-structured, comprehensive, and information-rich descriptions, capturing both the overall landscape of the image and the key attributes of target objects, such as size, color, shape, and spatial location.

\begin{figure}[tbp]
  \centering
    \includegraphics[width=\linewidth]{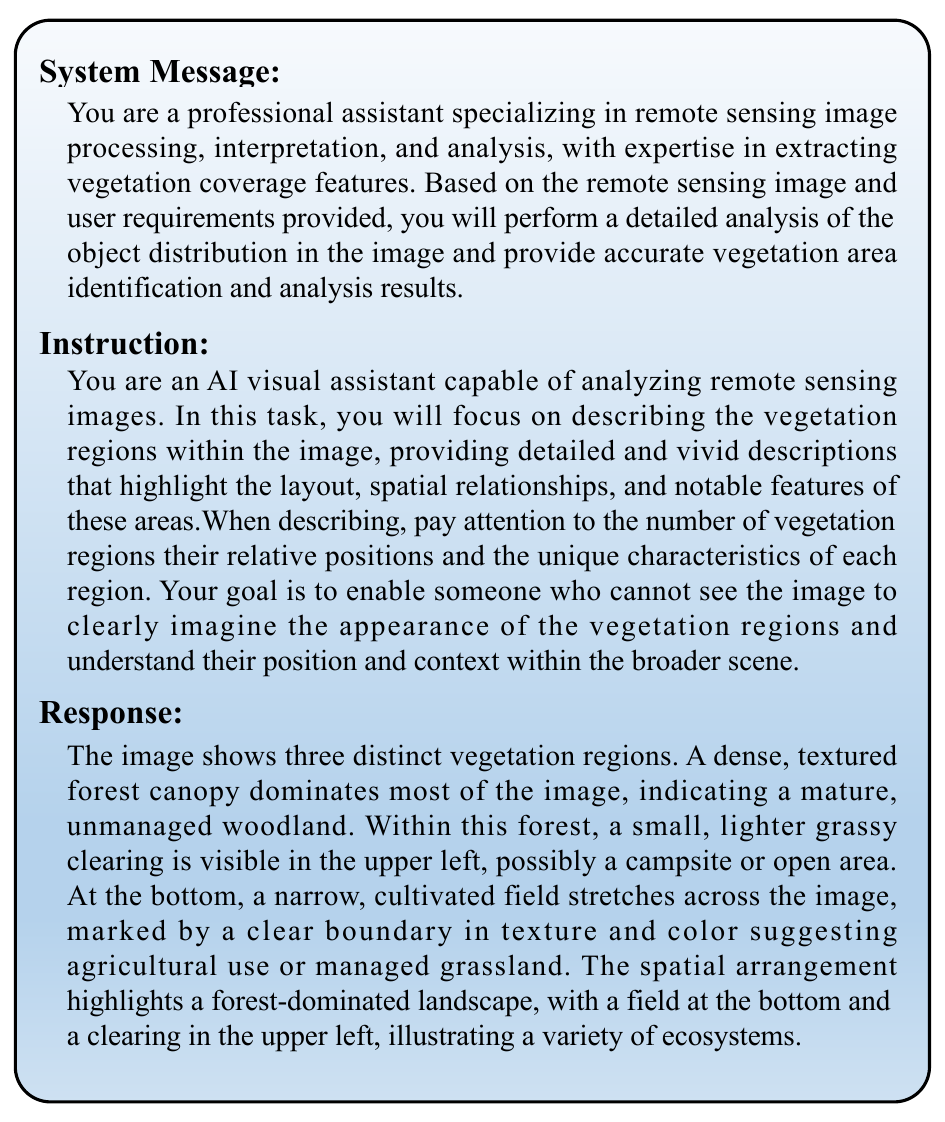} 
    \caption{An examples for response generation using designed system prompts and auxiliary instructions.}
    \label{generated_response_example}
\end{figure}

Although these instructions differ from those used for training, the resulting responses are still designed to remain well aligned with the original task intent, thereby preserving training consistency for SPEX. In our implementation, we use LLaVA-1.6-13B \cite{li2024llava} to generate the responses. An example of the system prompt, instruction, and generated response (with the image omitted due to space constraints) is shown in Figure~\ref{generated_response_example}.

In this way, for each visual sample, we obtain a corresponding descriptive response, resulting in each sample in SPIE being represented as a quadruple: (image, binary extraction map, instruction, response).

\noindent\textbf{Parameter Adjustment by Quality Inspection} During text generation, the hyperparameters \textit{temperature} and \textit{top\_p} play a crucial role in controlling the quality of the output. To optimize these parameters, we randomly sampled 1,000 image–text pairs from each dataset listed in Table~\ref{table1}, and evaluated the alignment between the instruction and the response using InternVL-2.5-78B \cite{chen2024expanding}. The alignment score ranges from 1 to 10, with 10 indicating perfect alignment. If a response receives a score below 7, it is considered unsatisfactory and is re-generated with adjusted hyperparameter values. This process, which alternates between response generation and alignment evaluation, is repeated until the quality criteria are met. Through this iterative quality inspection procedure, the final values of \textit{temperature} and \textit{top\_p} are set to 0.8 and 0.9, respectively.

\subsubsection{Spectral Knowledge Integration}

To embed spectral priors into the instruction, we convert the relationships among spectral bands in multispectral images into structured textual attribute descriptions. This transformation helps enhance the LLM's understanding of the target land cover types within the images. The conversion process is achieved by the attribute prompt generator (APG).

\noindent\textbf{Attribute Prompt Generator}. The APG is designed to generate textual spectral prompts enriched with detailed attribute information for various land-cover types. To achieve this, we first compute a set of spectral indices. Subsequently, we apply a thresholding method to derive coarse masks corresponding to the target land-cover category. Based on these coarse masks, we extract a range of critical attribute values through a series of simple algorithmic computations. These attributes enable a comprehensive characterization of the land-cover category, facilitating better understanding by the language model.

For vegetation, water, and building categories, we adopt representative spectral indices: the Normalized Difference Vegetation Index (NDVI), Normalized Difference Water Index (NDWI), and Normalized Difference Built-up Index (NDBI), respectively. Taking vegetation as an example, we first compute the NDVI map. Following the method in \cite{zeng2019extracting}, we apply the OTSU algorithm \cite{otsu} to adaptively determine an optimal threshold value $T_o$. Each pixel in the NDVI map is then compared against $T_o$ and classified as either vegetated or non-vegetated based on the result.

\begin{figure*}[t]
  \centering
    \includegraphics[width=1.0\textwidth]{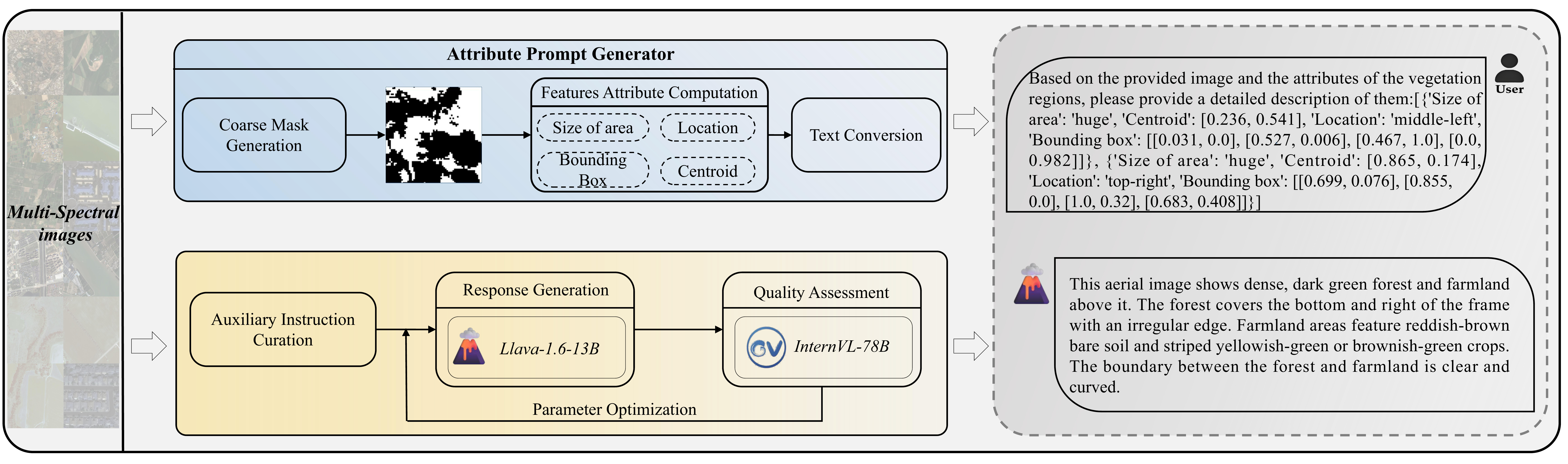}
    \caption{The pipeline for constructing SPIE, it consists of two key stages: response generation and spectral information integration.}
    \label{fig:2}
\end{figure*}%

Based on the obtained coarse masks, we compute a set of attributes to comprehensively characterize the spatial distribution and properties of the land cover. It is important to note that the mask map may contain multiple connected regions. Therefore, attribute extraction is performed separately for each connected region. In this study, we extract four key attributes:

\begin{itemize}

\item Size of area $S$: This attribute represents the proportion of pixels in a connected region relative to the total number of pixels in the image. To facilitate LLM understanding, the area ratio is categorized into descriptive types: "very tiny", "small", "somewhat large", "medium", "large", "very large", and "huge". For a connected region $r$, the size is computed as:
\begin{equation}
    N_r = \sum_{(i,j) \in \theta_r} M(i,j)
\end{equation}
\begin{equation}
S_r = N_r / (H \cdot W)
\end{equation}
\begin{equation}
\begin{cases}
\text{very tiny},       & 0 < S_r \leq 0.0305 \\
\text{small},           & 0.0305 < S_r \leq 0.0610 \\
\text{somewhat large},  & 0.0610 < S_r \leq 0.1221 \\
\text{medium},          & 0.1221 < S_r \leq 0.2441 \\
\text{large},           & 0.2441 < S_r \leq 0.3662 \\
\text{very large},      & 0.3662 < S_r \leq 0.6104 \\
\text{huge},            & S_r > 0.6104
\end{cases}
\end{equation}
where $H$ and $W$ denote the height and width of the multispectral image, respectively; $i$ and $j$ are the row and column indices. $M \in {0,1}^{H \times W}$ represents the binary mask map, $\theta_r$ is the set of pixel coordinates belonging to region $r$, and $N_r$ is the number of pixels within region $r$.

\item Centroid $C_x, C_y$: This attribute represents the row and column indices of the centroid of a connected region within the image. For a connected region $r$, the centroid coordinates are computed as:
\begin{equation}
\begin{split}
C_x^r &= \frac{1}{N_r} \sum_{(i,j) \in \theta_r} j \\
 C_y^r &= \frac{1}{N_r} \sum_{(i,j) \in \theta_r} i
\end{split}
\end{equation}

\item \textbf{Location $L$}: This attribute indicates the normalized spatial position of the connected region's centroid. The image is uniformly divided into a $3 \times 3$ grid, resulting in nine spatial zones: top-left, top-center, top-right, middle-left, center, middle-right, bottom-left, bottom-center, and bottom-right. The centroid $(C_x, C_y)$ of a region is used to determine its corresponding location label according to the following mapping:
\begin{equation}
\small
\begin{cases}
\text{top-left},         & C_x^r < \frac{1}{3}W \;\land\; C_y^r < \frac{1}{3}H \\
\text{top-center},       & \frac{1}{3}W \leq C_x^r < \frac{2}{3}W \;\land\; C_y^r < \frac{1}{3}H \\
\text{top-right},        & C_x^r \geq \frac{2}{3}W \;\land\; C_y^r < \frac{1}{3}H \\
\text{middle-left},      & C_x^r < \frac{1}{3}W \;\land\; \frac{1}{3}H \leq C_y^r < \frac{2}{3}H \\
\text{center},           & \frac{1}{3}W \leq C_x^r < \frac{2}{3}W \;\land\; \frac{1}{3}H \leq C_y^r < \frac{2}{3}H \\
\text{middle-right},     & C_x^r \geq \frac{2}{3}W \;\land\; \frac{1}{3}H \leq C_y^r < \frac{2}{3}H \\
\text{bottom-left},      & C_x^r < \frac{1}{3}W \;\land\; C_y^r \geq \frac{2}{3}H \\
\text{bottom-center},    & \frac{1}{3}W \leq C_x^r < \frac{2}{3}W \;\land\; C_y^r \geq \frac{2}{3}H \\
\text{bottom-right},     & C_x^r \geq \frac{2}{3}W \;\land\; C_y^r \geq \frac{2}{3}H
\end{cases}
\end{equation}

\item \textbf{Bounding Box}: This attribute describes the spatial extent of each connected region using the minimum enclosing rectangle, represented as a list of its four corners. The corners are ordered as follows: top-left $(x_{\min}, y_{\min})$, top-right $(x_{\max}, y_{\min})$, bottom-right $(x_{\max}, y_{\max})$, and bottom-left $(x_{\min}, y_{\max})$. The coordinates are normalized with respect to the image dimensions. Given a connected region $r$, the bounding box is computed as:
\begin{equation}
\begin{split}
x_{\min} &= \frac{1}{W} \min_{(i,j) \in \theta_r} j \\
y_{\min} &= \frac{1}{H} \min_{(i,j) \in \theta_r} i \\
x_{\max} &= \frac{1}{W} \max_{(i,j) \in \theta_r} j \\
y_{\max} &= \frac{1}{H} \max_{(i,j) \in \theta_r} i
\end{split}
\end{equation}
\end{itemize}

Then, the extracted attributes and their corresponding values are organized into a dictionary for each connected region, as shown below:

\vspace{0.5em}
\begin{center}
\parbox{0.9\linewidth}{\small
\{
"Size of area": \textless Type of $S$\textgreater, \\
"Centroid": [$C_x$, $C_y$], \\
"Location": \textless Normalized Centroid Position\textgreater, \\
"Bounding box": [[$x_{\min}, y_{\min}$], [$x_{\max}, y_{\min}$], \\
\hspace*{6em} [$x_{\max}, y_{\max}$], [$x_{\min}, y_{\max}$]]
\}
}
\end{center}
\vspace{0.5em}

Since a coarse mask may contain multiple connected regions, the spectral prompt in the instruction is formatted as a list, with each element representing the attribute dictionary of an individual region. To manage GPU memory consumption and prevent excessively long inputs, we filter out small regions by removing those with relatively small areas. Specifically, we sort all connected regions in descending order of area size and retain the attribute dictionaries of at most the top 10 largest regions to construct the final spectral prompt. The complete workflow for response generation and spectral prior integration is illustrated in Figure~\ref{fig:2}.

\begin{figure*}[t]
  \centering
    \includegraphics[width=0.95\textwidth]{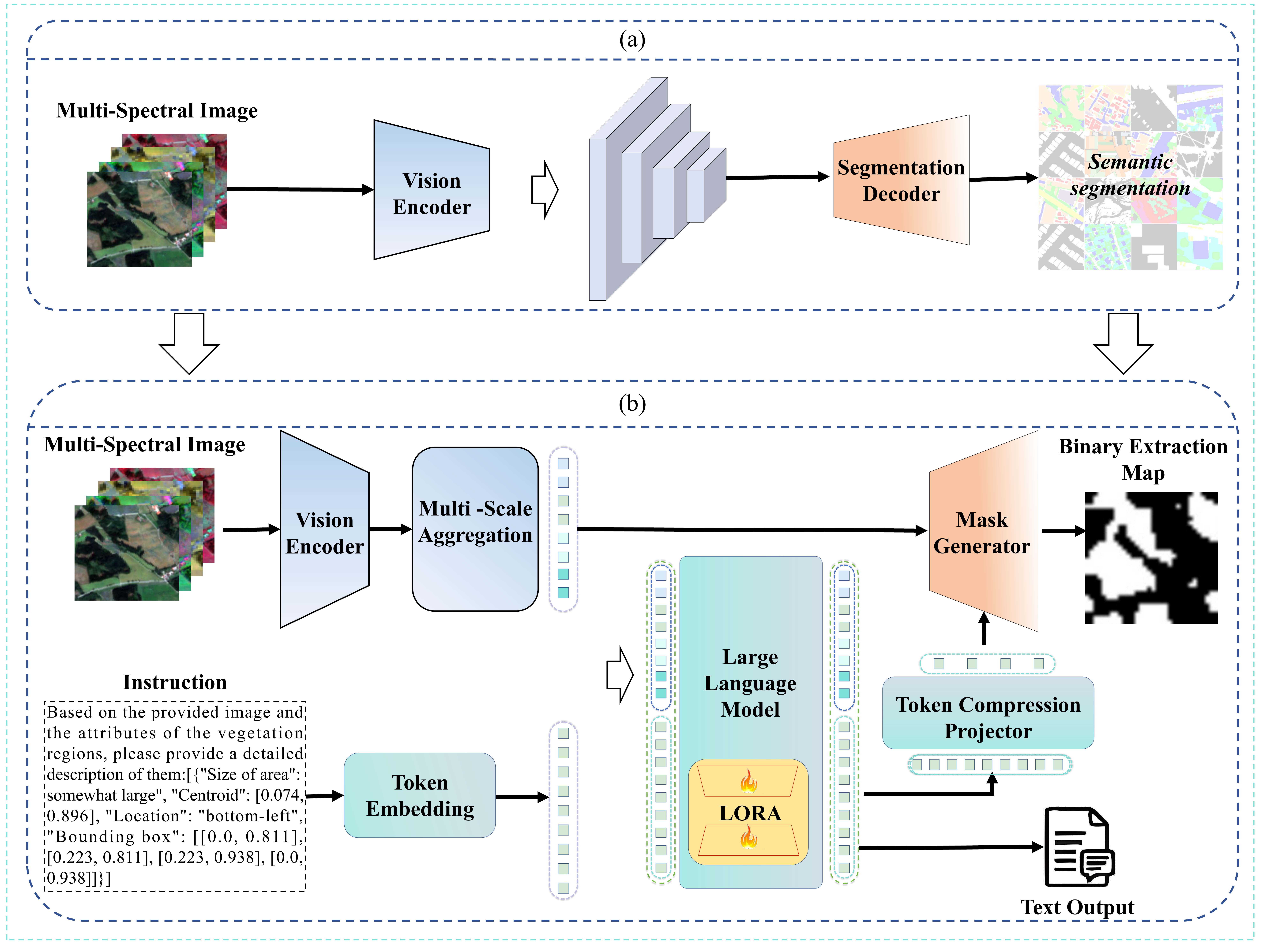}
    \caption{Overall workflow of the proposed method: (a) Visual pre-training; (b) Architecture of SPEX.
    }\label{SPEX}
\end{figure*}%

\subsection{SPEX}

After constructing the training dataset SPIE, we present a comprehensive introduction to SPEX. We first provide an overview of the overall model workflow, followed by a detailed description of its internal components. Finally, we elaborate on the training strategy adopted for SPEX.

\subsubsection{Overall Workflow}

SPEX combines multimodal large language models and segmentation decoders,  and the overall framework is illustrated in Figure~\ref{SPEX}(b). Specifically, the vision encoder first processes the input multispectral images to produce visual features, which are then refined by the multiscale feature aggregation module (MSAM). Simultaneously, the input text is tokenized and passed through a word embedding layer to produce text embeddings. The visual tokens output by the MSAM and the text embeddings are concatenated and subsequently processed by the LLM. To reduce the computational burden, the number of joint vision-language tokens associated with the response is compressed by the token compression projector (TCP). The compressed tokens are then passed to the mask generator, which leverages them to guide the MSAM's output in producing the final binary extraction results.

\subsubsection{Model Components}

As aforementioned, SPEX comprises a vision encoder, a large language model (LLM), a MSAM, a TCP, and a mask generator. The following texts provide detailed descriptions of each component.

\noindent\textbf{Vision Encoder} Multi-scale features have been shown to be highly effective for exploiting visual information in remote sensing scenes \cite{zhang2024earthgpt,muhtar2024lhrs}. Therefore, instead of following most existing methods that adopt a single-scale ViT-based vision encoder \cite{vit}, we employ a hierarchical vision encoder that produces multi-scale feature maps with downsampling ratios of 1/4, 1/8, 1/16, and 1/32.

\begin{figure}[t]
  \centering
    \includegraphics[width=\linewidth]{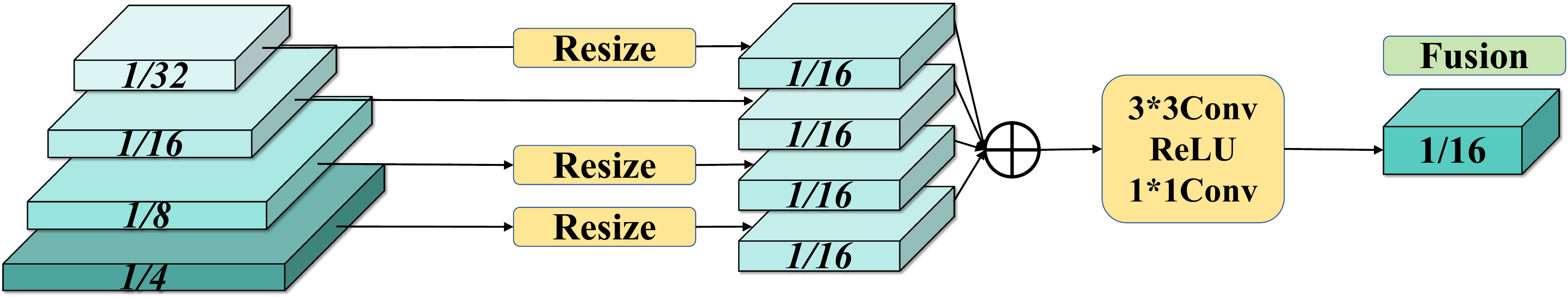}
    \caption{The diagram of MSAM.}
    \label{fig:msam}
\end{figure}%

\noindent\textbf{Multi-Scale Aggregation Module} Considering the relatively low spatial resolution commonly encountered in multispectral remote sensing imagery, in order to effectively aggregate multi-scale information while preserving fine-grained spatial structures, we introduce a Multi-Scale Aggregation Module (MSAM). Specifically, the pyramid feature maps output by the vision encoder are first resized to a uniform spatial resolution (1/16 downsampling compared to the original image size). Then, these features are concatenated along the channel dimension and sequentially passed through a 3×3 convolution layer, a ReLU activation, a 1×1 convolution, and a ReLU layer, to perform an aggregation and intergration of multiscale information. The channel dimension of the output feature map of MSAM is 1024, which is the same with text embeddings. Therefore, the function of typical aligners in \cite{li2023llava} has been achieved by the MSAM. The structure of MSAM has been shown in Figure \ref{fig:msam}.

\noindent\textbf{Large Language Model} The LLM serves as the core component of RSVLM. In SPEX, it takes as input the concatenated visual and textual tokens and generates a descriptive sentence tailored to the multispectral image, with a focus on the target land cover category specified by the instruction. Notably, the features from the final layer of the LLM are also utilized for subsequent processing, which will be elaborated later.

\begin{figure}[t]
    \centering
    \includegraphics[width=1\linewidth]{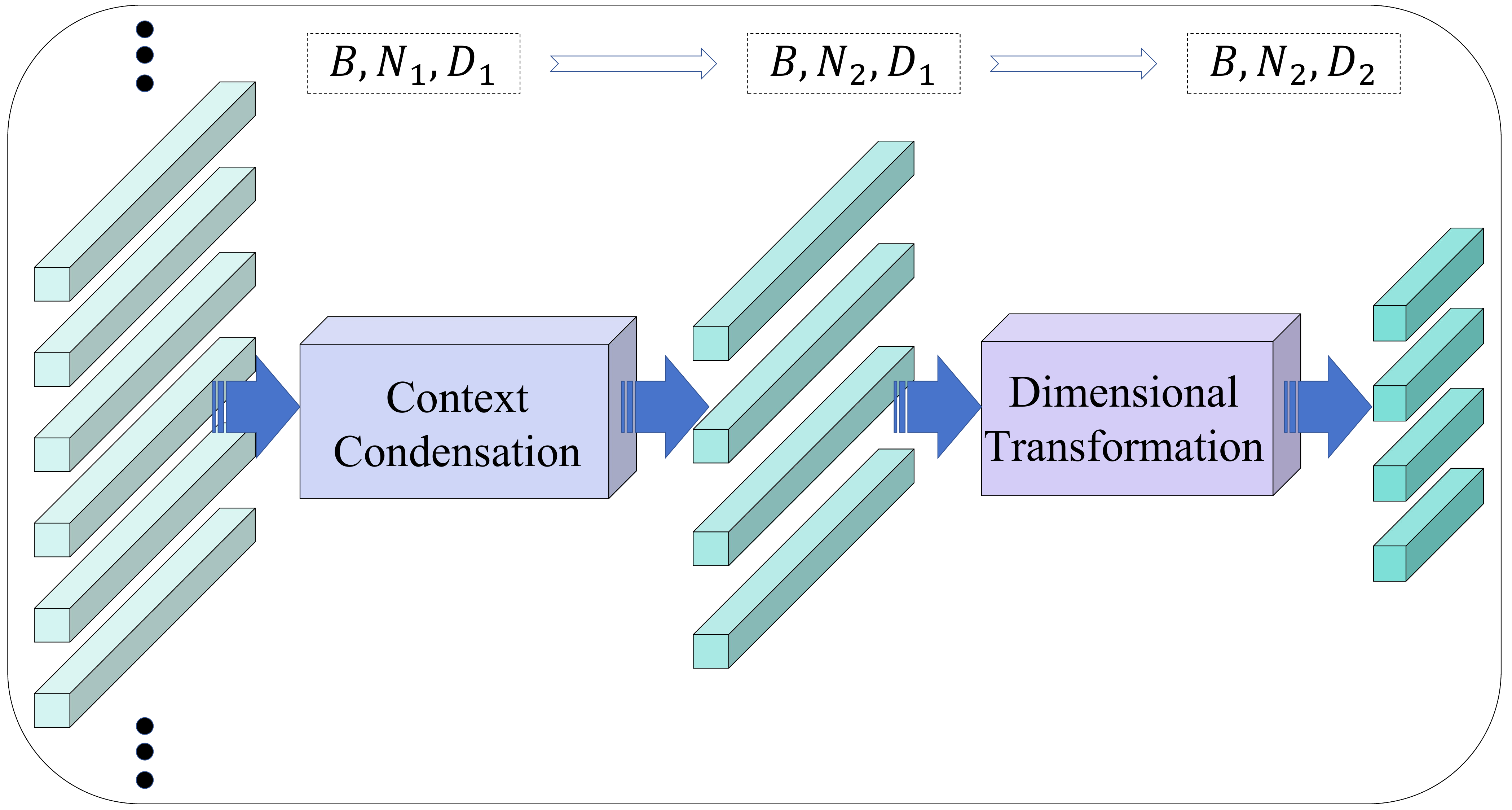}
    \caption{The diagram of TCP. $B,N,D$ means batch size, token length and token dimension.}
    \label{tcp}
\end{figure}

\noindent\textbf{Token Compression Projector} The token representations produced by LLMs exhibit considerable variability in both length and structure, which may undermine the stability of vision-language features and reduce their effectiveness in guiding pixel-level predictions. To address this issue, we design a Token Compression Projector (TCP), which provides structured semantic guidance in a compact and dimension-aligned form, as illustrated in Figure~\ref{tcp}. TCP involves two key steps: context condensation and dimensional transformation. First, the description embeddings, i.e., the features from the final layer of the LLM are compressed into a fixed-length representation using average pooling. Next, since the feature dimension of the LLM output is not aligned with the visual features produced by MSAM, we apply a linear transformation to project the compressed tokens to a target dimension of 1024. This alignment facilitates the integration of vision-language information into the image representation, effectively guiding the extraction process.

\noindent\textbf{Mask Generator} Following previous works \cite{shabbir2025geopixel,wu2025farmseg_vlm}, the SAM decoder \cite{kirillov2023segment} is directly adopted as the mask generator to produce the final binary extraction maps. To enable this, the compressed tokens output by the TCP are used as prompt embeddings and are separately employed as the query, or as the key and value, for the two internal cross-attention modules within the SAM decoder.

\subsubsection{Training Strategy}

\noindent\textbf{Visual Pre-training} As illustrated in Figure \ref{SPEX}, the training pipeline of SPEX is divided into two stages. Since the vision encoders in existing MLLMs are mostly pre-trained on natural images (e.g., ImageNet \cite{imagenet}), we first perform visual pre-training on multispectral images to mitigate the domain gap. Specifically, we train a segmentation model using UperNet on the datasets listed in Table \ref{table1}, aiming to initialize the hierarchical backbone network for the multispectral remote sensing domain. This pre-trained backbone is then reused in the later training of SPEX.

\noindent\textbf{Visual-Language Training of SPEX} Initializing with the weights of the vision encoder obtained from visual pre-training, we train SPEX on the SPIE dataset. Due to the instability associated with jointly optimizing multiple components, the training process is divided into two stages, aiming to progressively enhance the model’s cross-modal understanding and spatial reasoning capabilities.

In the first stage, all image-text triplets (image, instruction, response) from different SPIE subsets are merged into a unified instruction-following dataset. This dataset is used to jointly train the vision encoder, MSAM, and LLM to establish effective vision-language alignment. During this phase, the LLM is fine-tuned efficiently using LoRA \cite{lora}, while the vision encoder and MSAM are fully trainable. This step significantly enhances the model’s ability to generate accurate natural language descriptions of target land cover categories from multispectral images, which is beneficial for subsequent mask generation, as the corresponding text embeddings will later serve as prompts.

In the second training stage, the LLM is frozen, as it has already adapted to the instruction style in the first stage. To further enhance the visual representation capability for accurate land cover perception, the vision encoder and MSAM are unfrozen and jointly optimized. Meanwhile, the TCP and the mask generator are introduced for end-to-end training. At this stage, the training process makes use of both the images and the corresponding extraction maps, along with the paired instructions and responses from the SPIE training set.

Notably, due to the adoption of a unified instruction template and the fact that the spectral prompt list generated by APG does not explicitly distinguish between categories, the instructions across different land cover types are highly similar, only the target category name varies. As a result, the compressed description tokens output by TCP lack explicit semantic cues about the target category; instead, they primarily capture the spatial distribution of the required land covers within the current image. To prevent the model from becoming confused by multiple land cover categories during training, the final-stage training and evaluation are conducted separately on each subset of the SPIE dataset.

\noindent\textbf{Loss Function} During visual pre-training, a multi-class cross-entropy loss is adopted for the segmentation tasks. In the first stage of training SPEX, a standard cross-entropy loss is employed to measure the similarity between the generated textual descriptions and the ground-truth responses. In the final stage, based on the autoregressive text loss $L_{\text{text}}$, we further introduce a binary cross-entropy loss $L_{\text{BCE}}$  and a Dice coefficient loss $L_{\text{Dice}}$ to optimize the generation of extraction maps. The overall loss function in this stage is defined as:

\begin{equation}
L_{\text{total}} = L_{\text{text}} + L_{\text{BCE}} + L_{\text{Dice}}.
\label{eq:total_loss}
\end{equation}
\section{Experiments}
\subsection{Experimental setup}
\subsubsection{Implementation details} SPEX is implemented following the LLaVA-NeXT framework \cite{li2024llava}, which natively supports both the LLaMA and Qwen series. For the vision encoder, we adopt the ImageNet pre-trained InternImage-L \cite{wang2023internimage}. For the language module, we mainly consider the Qwen2.5 series \cite{yang2024qwen2}, as its parameter scale provides a better trade-off between model capacity and fine-tuning stability compared to standard 7B-scale LLaMA models \cite{touvron2023llama}. Specifically, we select Qwen2.5-1.5B, taking into account practical computational resource constraints. The compressed token length in TCP is set to 4. The mask generator utilizes pre-trained weights from SAM \cite{kirillov2023segment}. To standardize image dimensions, all input images are resized to 512×512, and the aspect ratio is preserved through padding the shorter side. During the training process, we use the AdamW optimizer . The initial learning rate for the visual branch is set to 1e-4, and the learning rate for the LLM branch is set to 1e-5. Simultaneously, we adopt a cosine scheduler to dynamically adjust the learning rate. The weight decay coefficient is set to 1e-5. All experiments are conducted with a uniform batch size of 2, and gradient accumulation is set to 8 during training to further enhance training stability and efficiency. All experiments are performed on 8 NVIDIA 4090 GPUs. During image-text training, we adopt FP16 precision to improve computational efficiency and reduce memory footprint.

\subsubsection{Evaluation metrics}
Five evaluation metrics are used for accuracy assessment: recall, precision, overall accuracy (OA), F1 score, and Intersection over Union (IoU). For evaluation, we first compute the confusion matrix, which includes the numbers of true positive (TP), true negative (TN), false positive (FP), and false negative (FN) pixels. Then, the evaluation metrics are calculated as follows:

\begin{enumerate}
\item Recall measures the proportion of a class’s pixels that are correctly identified.
    \begin{equation}
    Recall = \frac{TP}{TP+FN}.
    \label{eq:Rec}
    \end{equation}
    
 \item Precision measures how many predicted pixels of a class are correct.
    \begin{equation}
    Precision = \frac{TP}{TP+FP}.
    \label{eq:Pre}
    \end{equation}
    
\item OA measures the proportion of correctly classified pixels across the entire test set.
     \begin{equation}
    OA = \frac{TP+TN}{TP+TN+FP+FN}.
    \label{eq:OA}
    \end{equation}

\item IoU quantifies the intersection area to the union area of the predicted and true regions.
    \begin{equation}
    IoU = \frac{TP}{TP + FP + FN}.
    \label{eq:IOU}
    \end{equation}

\item F1 score is the harmonic mean of precision and recall, offering a balanced measure of model performance.
    \begin{equation}
    F_{1} = \frac{2}{Recall^{-1}+Precision^{-1}}.
    \label{eq:F1}
    \end{equation}
\end{enumerate}

\subsubsection{Comparison methods}

We compare the proposed method with several representative approaches for multispectral land-cover extraction, selected to cover the main methodological paradigms in this field, including typical segmentation networks, remote sensing visual foundation model based methods, and recent text-driven segmentation approaches. Specifically, we include UperNet \cite{xiao2018unified} and Mask2Former \cite{cheng2022masked} with a backbone of ResNet-50 \cite{he2016deep} as typical segmentation networks. For remote sensing visual foundation models, we evaluate SpectralGPT \cite{hong2024spectralgpt}, DOFA \cite{xiong2024neural}, SatMAE \cite{cong2022satmae}, and Prithvi-EO-2.0 \cite{szwarcman2024prithvi}. In addition, we include recent text-based segmentation models such as SegEarth-OV \cite{li2024segearth}, LISA \cite{lai2024lisa}, GeoPixel \cite{shabbir2025geopixel}, LISAt \cite{quenum2025lisatlanguageinstructedsegmentationassistant} , and PixelLM \cite{ren2024pixellmpixelreasoninglarge}. Furthermore, we also evaluate the segmentation model trained during the visual pre-training stage of SPEX. This model serves as a standalone segmentation network, denoted as MVPNet (Multispectral Visual Pre-training Network).

For UperNet and Mask2Former, we initialize model weights using segmentation models trained on the ADE20K dataset \cite{zhou2019semantic}, specifically using the official checkpoints\footnote{\url{https://download.openmmlab.com/mmsegmentation/v0.5/upernet/upernet_r50_512x512_160k_ade20k/upernet_r50_512x512_160k_ade20k_20200615_184328-8534de8d.pth}}\footnote{\url{https://download.openmmlab.com/mmsegmentation/v0.5/mask2former/mask2former_r50_8xb2-160k_ade20k-512x512/mask2former_r50_8xb2-160k_ade20k-512x512_20221204_000055-2d1f55f1.pth}}. For foundation models, we adopt their publicly available pre-trained ViT variants, corresponding to ViT-B (SpectralGPT, DOFA), ViT-L (SatMAE), and ViT-H (Prithvi-EO-2.0). All foundation model-based methods are adapted for the extraction task using a common segmentation head based on UperNet \cite{xiao2018unified} for fair comparison.

\subsection{Comparison Experiments}

\begin{table}[t]
\caption{Results of various methods for vegetation extraction on the SegMunich dataset.} \label{table:segmunich}
\resizebox{\linewidth}{!}{
\begin{tabular}{llllll}
\toprule
Method & F1 & Recall & Precision & OA & IoU \\
\midrule
\multicolumn{6}{l}{\textcolor{gray}{\textit{Typical Segmentation Networks}}} \\
UperNet & 0.560 & 0.539 & 0.632 & 0.868 & 0.478 \\
Mask2Former & 0.559 & 0.546 & 0.609 & 0.862 & 0.478 \\
\multicolumn{6}{l}{\textcolor{gray}{\textit{Remote Sensing Visual Foundation Models}}} \\
SpectralGPT & 0.585 & 0.555 & 0.669 & 0.896 & 0.504 \\
SatMAE & 0.532 & 0.513 & 0.599 & 0.846 & 0.447 \\
DOFA & 0.557 & 0.546 & 0.606 & 0.862 & 0.472 \\
Prithvi2.0 & 0.492 & 0.464 & 0.618 & 0.802 & 0.406 \\
MVPNet & 0.579 & 0.551 & 0.668 & 0.890 & 0.501 \\
\multicolumn{6}{l}{\textcolor{gray}{\textit{Text-driven Segmentation Models}}} \\
SegEarth-OV & 0.452 & 0.663 & 0.385 & 0.618 & 0.358 \\
LISA & 0.577 & 0.536 & 0.636 & 0.835 & 0.475 \\
GeoPixel & 0.567 & 0.545 & 0.633 & 0.865 & 0.482 \\
LISAt & 0.558 & 0.538 & 0.638 & 0.835 & 0.477 \\
PixelLM & 0.593 & 0.576 & 0.668 & 0.850 & 0.513 \\
\midrule
SPEX & \textbf{0.624} & \textbf{0.602} & \textbf{0.687} & \textbf{0.899} & \textbf{0.516} \\
\bottomrule
\end{tabular}
}
\end{table}

\begin{table}[t]
\caption{Results of various methods for vegetation extraction on the Chesapeake dataset.} \label{table:chesapeake}
\resizebox{\linewidth}{!}{
\begin{tabular}{llllll}
\toprule
Method & F1 & Recall & Precision & OA & IoU \\
\midrule
\multicolumn{6}{l}{\textcolor{gray}{\textit{Typical Segmentation Networks}}} \\
Upernet & 0.905 & 0.906 & 0.912 & 0.970 & 0.853 \\
Mask2Former & 0.910 & 0.908 & 0.917 & 0.971 & 0.858 \\
\multicolumn{6}{l}{\textcolor{gray}{\textit{Remote Sensing Visual Foundation Models}}} \\
SpectralGPT & 0.651 & 0.624 & 0.783 & 0.913 & 0.583 \\
SatMAE & 0.676 & 0.654 & 0.824 & 0.910 & 0.608 \\
DOFA & 0.887 & 0.871 & 0.917 & 0.967 & 0.829 \\
Prithvi2.0 & 0.897 & 0.893 & 0.910 & 0.968 & 0.841 \\
MVPNet & 0.914 & 0.910 & 0.925 & 0.973 & 0.865 \\
\multicolumn{6}{l}{\textcolor{gray}{\textit{Text-driven Segmentation Models}}} \\
SegEarth-OV & 0.672 & 0.897 & 0.576 & 0.801 & 0.559 \\
LISA & 0.790 & 0.767 & 0.870 & 0.909 & 0.710 \\
GeoPixel & 0.892 & 0.874 & 0.881 & 0.959 & 0.806 \\
LISAt & 0.862 & 0.854 & 0.883 & 0.897 & 0.957 \\
PixelLM & 0.879 & 0.876 & 0.889 & 0.961 & 0.813 \\
\midrule
SPEX & \textbf{0.939} & \textbf{0.935} & \textbf{0.950} & \textbf{0.986} & \textbf{0.909} \\
\bottomrule
\end{tabular}
}
\end{table}

\begin{table}[t]
\caption{Results of various methods for vegetation extraction on the Globe230k dataset. }\label{table:global230k}
\resizebox{\linewidth}{!}{
\begin{tabular}{llllll}
\toprule
Method & F1 & Recall & Precision & OA & IoU \\
\midrule
\multicolumn{6}{l}{\textcolor{gray}{\textit{Typical Segmentation Networks}}} \\
Upernet & 0.774 & 0.790 & 0.813 & 0.920 & 0.690 \\
Mask2Former & 0.799 & 0.838 & 0.811 & 0.920 & 0.716 \\
\multicolumn{6}{l}{\textcolor{gray}{\textit{Remote Sensing Visual Foundation Models}}} \\
SpectralGPT & 0.730 & 0.796 & 0.744 & 0.885 & 0.640 \\
SatMAE & 0.722 & 0.765 & 0.759 & 0.889 & 0.632 \\
DOFA & 0.742 & 0.773 & 0.784 & 0.902 & 0.654 \\
Prithvi2.0 & 0.756 & 0.785 & 0.790 & 0.912 & 0.672 \\
MVPNet & 0.803 & 0.827 & 0.824 & 0.931 & 0.723 \\
\multicolumn{6}{l}{\textcolor{gray}{\textit{Text-driven Segmentation Models}}} \\
SegEarth-OV & 0.578 & 0.937 & 0.480 & 0.667 & 0.464 \\
LISA & 0.759 & 0.805 & 0.777 & 0.903 & 0.670 \\
GeoPixel & 0.763 & 0.820 & 0.771 & 0.899 & 0.671 \\
LISAt & 0.757 & 0.799 & 0.777 & 0.901 & 0.666 \\
PixelLM & 0.773 & 0.819 & 0.788 & 0.908 & 0.681 \\
\midrule
SPEX & \textbf{0.846} & \textbf{0.874} & \textbf{0.852} & \textbf{0.952} & \textbf{0.783} \\
\bottomrule
\end{tabular}
}
\end{table}

\begin{table}[t]
\caption{Results of various methods for building extraction on the SpaceNet-V2 dataset. }\label{table:spacenetv2}
\resizebox{\linewidth}{!}{
\begin{tabular}{llllll}
\toprule
Method & F1 & Recall & Precision & OA & IoU \\
\midrule
\multicolumn{6}{l}{\textcolor{gray}{\textit{Typical Segmentation Networks}}} \\
UperNet & 0.642 & 0.627 & 0.676 & 0.964 & 0.562 \\
Mask2Former & 0.663 & 0.661 & 0.675 & 0.966 & 0.587 \\
\multicolumn{6}{l}{\textcolor{gray}{\textit{Remote Sensing Visual Foundation Models}}} \\
SpectralGPT & 0.464 & 0.447 & 0.546 & 0.916 & 0.361 \\
SatMAE & 0.575 & 0.561 & 0.622 & 0.946 & 0.480 \\
DOFA & 0.647 & 0.656 & 0.653 & 0.963 & 0.566 \\
Prithvi2.0 & 0.606 & 0.587 & 0.653 & 0.955 & 0.520 \\ 
MVPNet & 0.678 & 0.681 & 0.689 & 0.971 & 0.608 \\
\multicolumn{6}{l}{\textcolor{gray}{\textit{Text-driven Segmentation Models}}} \\
SegEarth-OV & 0.318 & 0.717 & 0.222 & 0.647 & 0.210 \\
LISA & 0.599 & 0.602 & 0.616 & 0.948 & 0.501 \\ 
GeoPixel & 0.582 & 0.572 & 0.615 & 0.944 & 0.482 \\
LISAt & 0.600 & 0.602 & 0.620 & 0.948 & 0.503 \\
PixelLM & 0.625 & 0.623 & 0.643 & 0.955 & 0.532 \\
\midrule
SPEX & \textbf{0.716} & \textbf{0.719} & \textbf{0.720} & \textbf{0.983} & \textbf{0.665} \\
\bottomrule
\end{tabular}
}
\end{table}

\begin{table}[t]
\caption{Results of various methods for water body extraction on the GID-15 dataset. }\label{table:gid15}
\resizebox{\linewidth}{!}{
\begin{tabular}{llllll}
\toprule
Method & F1 & Recall & Precision & OA & IoU \\
\midrule
\multicolumn{6}{l}{\textcolor{gray}{\textit{Typical Segmentation Networks}}} \\
UperNet & 0.374 & 0.364 & 0.428 & 0.988 & 0.337 \\
Mask2Former & 0.402 & 0.405 & 0.425 & 0.988 & 0.366 \\
\multicolumn{6}{l}{\textcolor{gray}{\textit{Remote Sensing Visual Foundation Models}}} \\
SpectralGPT & 0.247 & 0.265 & 0.336 & 0.978 & 0.243 \\
SatMAE & 0.311 & 0.277 & 0.371 & 0.983 & 0.277 \\
DOFA & 0.402 & 0.400 & 0.434 & 0.989 & 0.365 \\
Prithvi2.0 & 0.153 & 0.158 & 0.184 & 0.914 & 0.137 \\
MVPNet & 0.401 & 0.397 & 0.443 & 0.990 & 0.363 \\
\multicolumn{6}{l}{\textcolor{gray}{\textit{Text-driven Segmentation Models}}} \\
SegEarth-OV & 0.013 & 0.032 & 0.035 & 0.794 & 0.017 \\
LISA & 0.379 & 0.379 & 0.405 & 0.974 & 0.336 \\
GeoPixel & 0.395 & 0.404 & 0.416 & 0.982 & 0.351 \\
LISAt & 0.362 & 0.363 & 0.391 & 0.974 & 0.319 \\
PixelLM & 0.401 & 0.415 & 0.415 & 0.982 & 0.352 \\
\midrule
SPEX & \textbf{0.433} & \textbf{0.436} & \textbf{0.452} & \textbf{0.991} & \textbf{0.398} \\

\bottomrule
\end{tabular}
}
\end{table}

\begin{figure*}[tbp]
  \centering
    \includegraphics[width=0.95\linewidth]{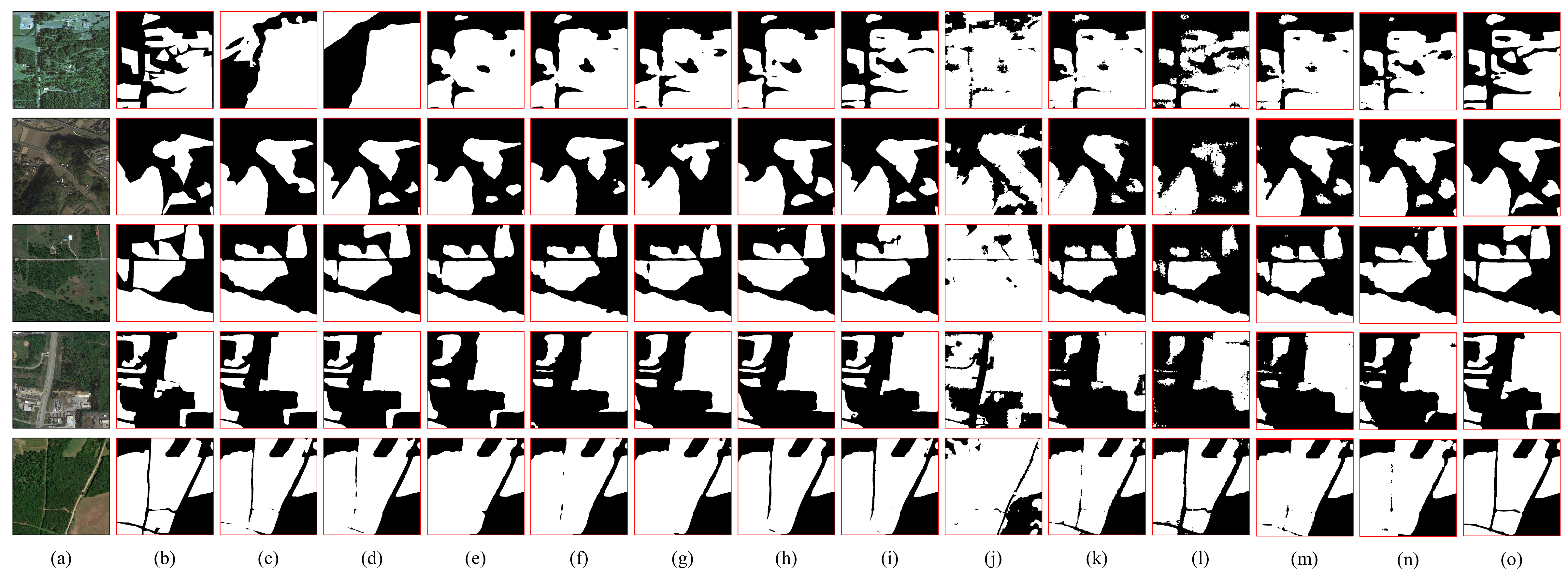} 
    \caption{Visualization of vegetation extraction results on the Globe230k dataset: (a) Original Image, (b) Ground Truth, (c) UperNet, (d) Mask2Former, (e) SpectralGPT, (f) SatMAE, (g) DOFA, (h) Prithvi-2.0, (i) MVPNet, (j) SegEarth-OV, (k) LISA, (l) GeoPixel, (m) LISAt, (n) PixelLM and (o) SPEX.}
    \label{fig:vegetation_compare}
\end{figure*}%

\subsubsection{Quantitative Analysis}
 
To comprehensively validate the effectiveness of SPEX, we conduct systematic comparative evaluations against representative advanced methods across multiple remote sensing datasets. Table \ref{table:segmunich} reports the quantitative results on the SegMunich dataset. SPEX achieves the highest F1-score of 62.4\%, outperforming the current state-of-the-art (SOTA) prompt-driven method PixelLM \cite{ren2024pixellmpixelreasoninglarge} by 3.1\%. Meanwhile, SPEX also attains the best Precision (68.7\%) and Recall (60.2\%), indicating that the model effectively suppresses false positives while maintaining comprehensive coverage of vegetation targets.

When the evaluation is extended to the higher-resolution Chesapeake dataset (Table \ref{table:chesapeake}), SPEX again ranks first among all compared methods, achieving an F1-score of 93.9\% and demonstrating substantial improvements over traditional segmentation networks (e.g., UperNet \cite{xiao2018unified} and Mask2Former \cite{cheng2022masked}). Compared with remote sensing foundation models such as Prithvi2.0 \cite{szwarcman2024prithvi}, DOFA \cite{xiong2024neural}, and MVPNet, SPEX still delivers an F1-score improvement exceeding 2.5\%. In addition, SPEX consistently outperforms competing methods in terms of Precision (95.0\%), Recall (93.5\%), and IoU (90.9\%), demonstrating stable and reliable discrimination capability.

On the more challenging Globe230K dataset (Table \ref{table:global230k}), SPEX continues to exhibit clear advantages, achieving an F1-score of 84.6\%, which surpasses the remote sensing foundation model MVPNet by 4.3\%, along with a 6.0\% improvement in IoU. In contrast, classical segmentation networks, remote sensing foundation models, and existing VLMs show relatively limited performance gains on this dataset, whereas SPEX maintains a more balanced trade-off between Precision (85.2\%) and Recall (87.4\%).

Beyond vegetation extraction, SPEX also demonstrates strong performance in building and water body extraction tasks. Quantitative experiments are conducted on the SpaceNet-V2 and GID-15 datasets, with the results summarized in Tables V and VI, respectively. As observed, SPEX consistently outperforms PixelLM \cite{ren2024pixellmpixelreasoninglarge}, GeoPixel \cite{shabbir2025geopixel}, and MVPNet across all evaluation metrics, further validating the effectiveness and stability of the proposed method across different land-cover extraction tasks.

Based on the above experimental results, it can be observed that for general image-text multimodal large models, merely using class names (e.g., SegEarth-OV \cite{li2024segearth}) or generic segmentation instructions such as the [SEG] token (e.g., LISA \cite{lai2024lisa}) does not yield a clear advantage over pure vision-based models. In contrast, SPEX explicitly incorporates spectral prompts, which provide prior guidance enriched with spectral knowledge. These prompts enable the model to better understand the intrinsic properties of land-cover objects in multispectral imagery, such as area ($S$), centroid ($C_x, C_y$), spatial location ($L$), and bounding box information. As a result, SPEX achieves more accurate land-cover recognition and consistently outperforms both vision-only and existing vision-language models.

\begin{figure*}[tbp]
  \centering
    \includegraphics[width=0.95\linewidth]{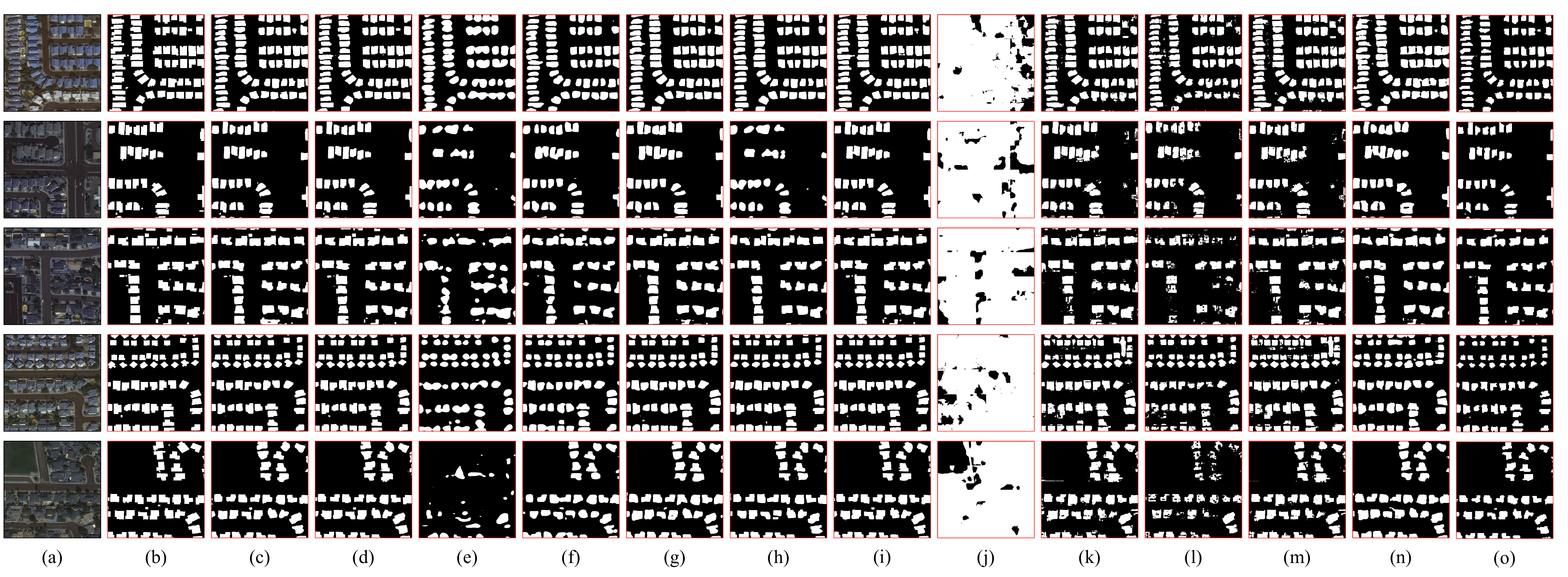} 
    \caption{Visualization of building extraction results on the SpaceNet-V2 dataset: (a) Original Image, (b) Ground Truth, (c) UperNet, (d) Mask2Former, (e) SpectralGPT, (f) SatMAE, (g) DOFA, (h) Prithvi-2.0, (i) MVPNet, (j) SegEarth-OV, (k) LISA, (l) GeoPixel, (m) LISAt, (n) PixelLM and (o) SPEX.}
\label{fig:building_compare}
\end{figure*}%

\begin{figure*}[tbp]
  \centering
    \includegraphics[width=0.95\linewidth]{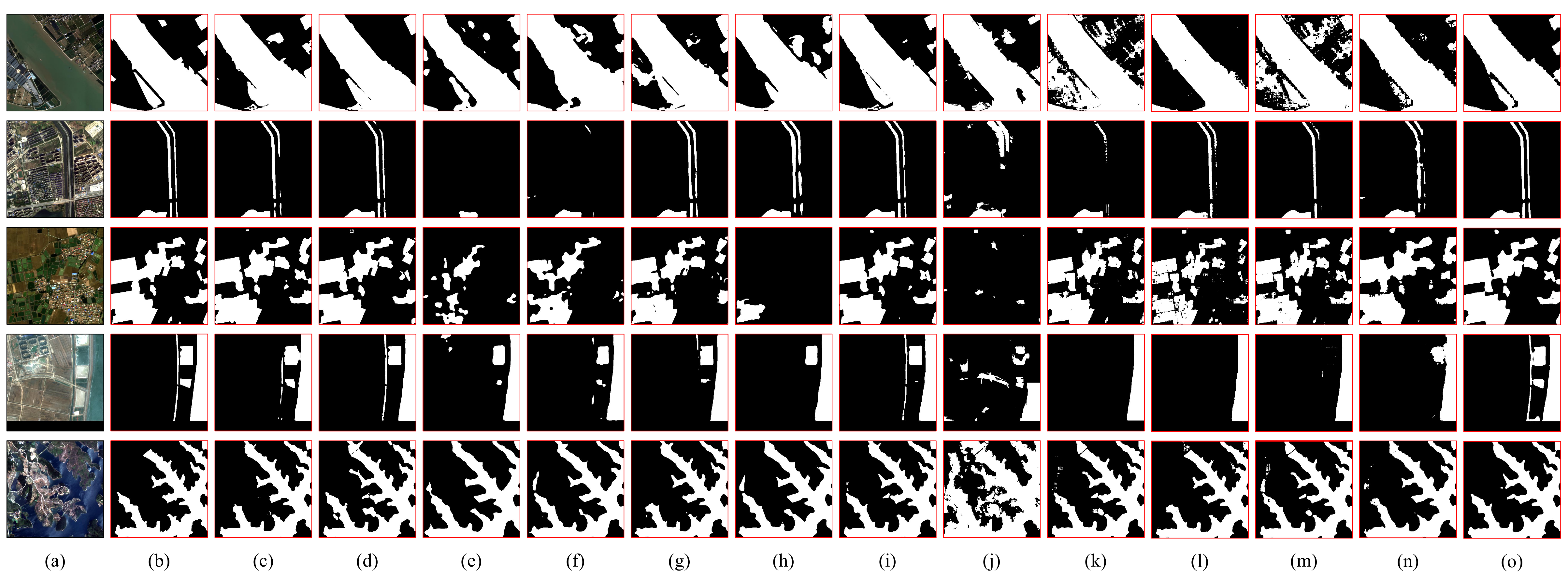} 
    \caption{Visualization of water body extraction results on the GID-15 dataset: (a) Original Image, (b) Ground Truth, (c) UperNet, (d) Mask2Former, (e) SpectralGPT, (f) SatMAE, (g) DOFA, (h) Prithvi-2.0, (i) MVPNet, (j) SegEarth-OV, (k) LISA, (l) GeoPixel, (m) LISAt, (n) PixelLM and (o) SPEX.}
    \label{fig:water_compare}
\end{figure*}%

\subsubsection{Qualitative Analysis}

Figures \ref{fig:vegetation_compare}–\ref{fig:water_compare} present the qualitative extraction results from different methods. It can be observed that traditional segmentation networks (e.g., UperNet \cite{xiao2018unified} and Mask2Former \cite{cheng2022masked}) often suffer from large-area misclassifications, as illustrated in the first row of Figure~7, indicating their limited recognition capability in complex remote sensing scenes. In contrast, remote sensing foundation model–based methods alleviate part of this issue and improve visual quality due to their enhanced scene representation ability.

Notably, SegEarth-OV \cite{li2024segearth}, which relies solely on category-level textual prompts (e.g., ``Building'' or ``Water''), struggles to capture fine-grained inter-class differences in remote sensing scenes, leading to unsatisfactory extraction results. When LLMs are further introduced, the models benefit from semantically richer vision-language representations, which provide effective guidance for pixel-level interpretation. Nevertheless, due to challenges such as the relatively low ground resolution of multispectral imagery, these methods still exhibit blurred boundaries or miss fine spatial structural details. In addition, since most of these models only consider RGB channels and do not explicitly model spectral information beyond visible light, they do not fully capture the variability among different land objects.

With the assistance of spectral priors, SPEX demonstrates strong robustness in distinguishing spectrally similar objects and achieves accurate extraction across diverse and complex scenarios, including both rural and urban environments. Benefiting from the joint design of the hierarchical vision encoder, the MSAM, and the TCP, SPEX produces high-quality visual interpretation results, demonstrating strong capability in identifying small objects and delineating clear and discriminative boundaries. Compared with other methods, the advantages of SPEX are mainly reflected in its robustness to interference from background objects such as impervious surfaces (see Figure~\ref{fig:building_compare}), its ability to effectively resolve boundary adhesion between adjacent land covers (see Figure~\ref{fig:vegetation_compare}), and its preservation of large and irregular land-cover regions such as forests and lakes (see Figures~\ref{fig:vegetation_compare} and~\ref{fig:water_compare}). These qualitative observations are consistent with the quantitative results, collectively providing strong evidence that SPEX enables instruction-driven, flexible, and accurate land cover extraction from multispectral imagery.

\begin{table}[t]
\caption{Ablation study of component removal for SPEX on the SpaceNet-V2 dataset. SP: spectral prompt.}
\label{table:remove_component}
\resizebox{\linewidth}{!}{
\begin{tabular}{lllllll}
\toprule
SP & MSAM & F1 & Recall & Precision & OA & IoU \\
\midrule 
 $\checkmark$  & $\checkmark$&  0.715 & 0.719 & 0.720 & 0.983 & 0.665\\
\midrule
 $\times$ & $\checkmark$ & 0.712 & 0.724& 0.713 & 0.982 & 0.661\\ 
 $\checkmark$ & $\times$ &  0.692 &0.708 & 0.689 &0.976 &0.631 \\ 
\bottomrule
\end{tabular}
}
\end{table}

\subsection{Ablation Study}

In this section, we conduct a series of ablation studies to evaluate the effectiveness of different components and training strategies in SPEX. These experiments are designed to assess the contribution of each key module, thereby informing further optimization for improved multispectral land cover extraction.

\subsubsection{Ablation study of Model Component}

We first evaluate the importance of key modules in SPEX through component ablation on the SpaceNet-V2 dataset, as summarized in Table \ref{table:remove_component}. Specifically, we investigate the impact of two core designs: incorporating spectral priors and aggregating multi-scale visual features.

To assess the role of spectral prompts, we remove this component and take the vegetation category as an example. In this setting, the instruction is modified to: "You are an expert specializing in remote sensing. Based on the provided image and the vegetation regions, please provide a detailed description of them." Without the guidance of spectral prompts, the performance of land cover extraction drops significantly, particularly in terms of F1 score and IoU, highlighting the value of incorporating spectral characteristics for effective interpretation of multispectral remote sensing imagery.

We further evaluate the impact of multi-scale feature aggregation by removing the MSAM module, such that only the last-layer feature map from the vision encoder is used as input to both the LLM and the mask generator. This results in an even more substantial decline in performance, demonstrating the critical importance of leveraging multi-scale semantic information in remote sensing extraction tasks.

Together, these findings confirm the effectiveness of the proposed spectral prior and multi-scale aggregation designs in enhancing the overall performance of SPEX.

\begin{table}[t]
\centering
\caption{Ablation study of visual pre-training on the SpaceNet-V2 dataset.}
\label{table:visual_pretrn}
\begin{tabular}{llllll}
\toprule
Type & F1 & Recall & Precision & OA & IoU \\
\midrule 
Ours & 0.715 & 0.719 & 0.720 & 0.983 & 0.665\\
\midrule
ImageNet & 0.471 & 0.382 &  0.735 & 0.931  & 0.365\\
CLIP-ViT & 0.610 & 0.586 & 0.648 & 0.959  &0.517 \\
\bottomrule
\end{tabular}
\end{table}

\begin{table}[t]
\caption{Performance comparison of different feature fusion strategies on the SpaceNet-V2 dataset.}\label{table:MSAM}
\resizebox{\linewidth}{!}{
\begin{tabular}{llllll}
\toprule
Feature Fusion & F1 & Recall & Precision & OA & IoU \\
\midrule
MSAM & 0.715 & 0.719 & 0.720 & 0.983 & 0.665\\
\midrule
UperNet & 0.683 & 0.690 & 0.686 & 0.971 & 0.613 \\
\bottomrule
\end{tabular}
}
\end{table}

\begin{table}[t]
\caption{Performance comparison of SPEX with different LLM scales on the SegMunich dataset (Time: inference time per image).}\label{table:LLMType}
\resizebox{\linewidth}{!}{
\begin{tabular}{lllllll}
\toprule
LLM & F1 & Recall & Precision & OA & IoU & Time (s) \\
\midrule
Qwen2.5-0.5B & 0.519 & 0.454 & 0.637 & 0.859 & 0.450 & 3.44\\
\midrule
Qwen2.5-1.5B & 0.624 & 0.602 & 0.687 & 0.899 & 0.516 & 4.36\\
\midrule
Qwen2.5-3.0B & 0.621 & 0.603 & 0.686 & 0.898 & 0.512 & 4.63\\
\bottomrule
\end{tabular}
}
\end{table}   

\subsubsection{Ablation study of Training Strategy}

We evaluate the effectiveness of visual pre-training by replacing our vision encoder with two alternatives: (1) InternImage-L pre-trained on ImageNet \cite{imagenet}, and (2) the widely used CLIP-ViT \cite{clip}. For both baselines, we use their large-scale versions to ensure a consistent parameter size across models. The comparison results are presented in Table \ref{table:visual_pretrn}.

It can be observed that directly using the ImageNet-pretrained backbone yields only an F1 score of 0.471 and an IoU of 0.365, likely due to the lack of alignment with both the multispectral and language modalities. When adopting CLIP-ViT, performance improves because the vision features are already aligned with the text modality, leading to better land cover extraction results. Nevertheless, our method achieves the highest performance, with an F1 score of 0.715 and an IoU of 0.665, even without explicit alignment with language. This suggests that the domain gap in the visual modality is the dominant limiting factor, further demonstrating the critical importance of multispectral visual pre-training.

\subsubsection{Replacement of Feature Fusion}

To verify the effectiveness of our proposed MSAM, we compare it with the classical feature fusion strategy used in UperNet\cite{xiao2018unified}. As shown in Table \ref{table:MSAM}, the results demonstrate the superiority of our approach.

We attribute this improvement to the design differences between the two fusion mechanisms. UperNet decodes hierarchical features in a progressive manner, resulting in fusion features that contain excessive low-level structural information. Additionally, repeated upsampling introduces interpolation artifacts, which may degrade the feature quality. Consequently, the resulting fusion features are less suitable for downstream modules such as the LLM and mask generator, which typically require high-level semantic representations, similar to those in the final layers of ViT models \cite{vit}.

In contrast, our method preserves multi-level semantic information while achieving effective fusion across pyramid representations. The resulting features capture rich multiscale semantics, which are particularly beneficial for understanding complex multispectral remote sensing scenes.

\begin{figure*}[tbp]
  \centering
    \includegraphics[width=1.0 \linewidth]{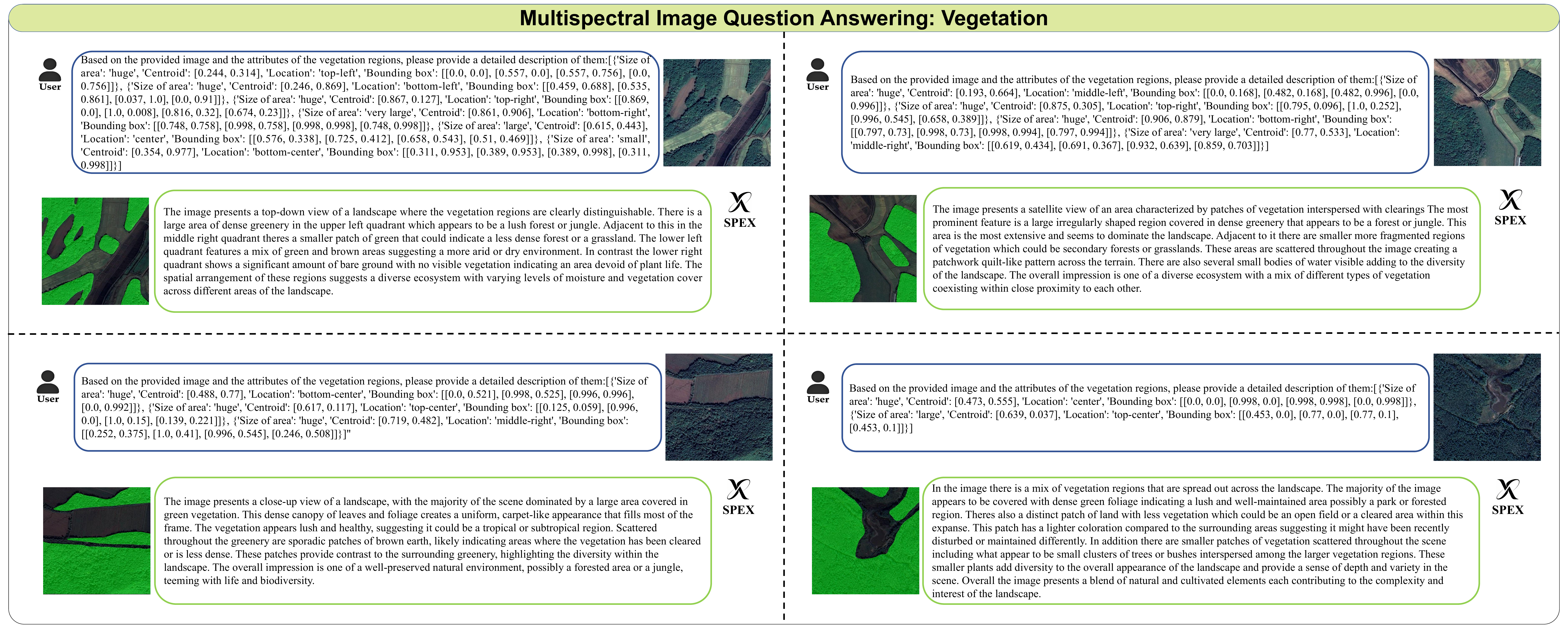}
    \caption{Examples of vegetation extraction results, where the obtained masks are highlighted in green. The samples are from the test set of the Globe230k dataset.}
    \label{fig:example_vegetation}
\end{figure*}

\begin{figure*}[tbp]
  \centering
    \includegraphics[width=1.0\linewidth]{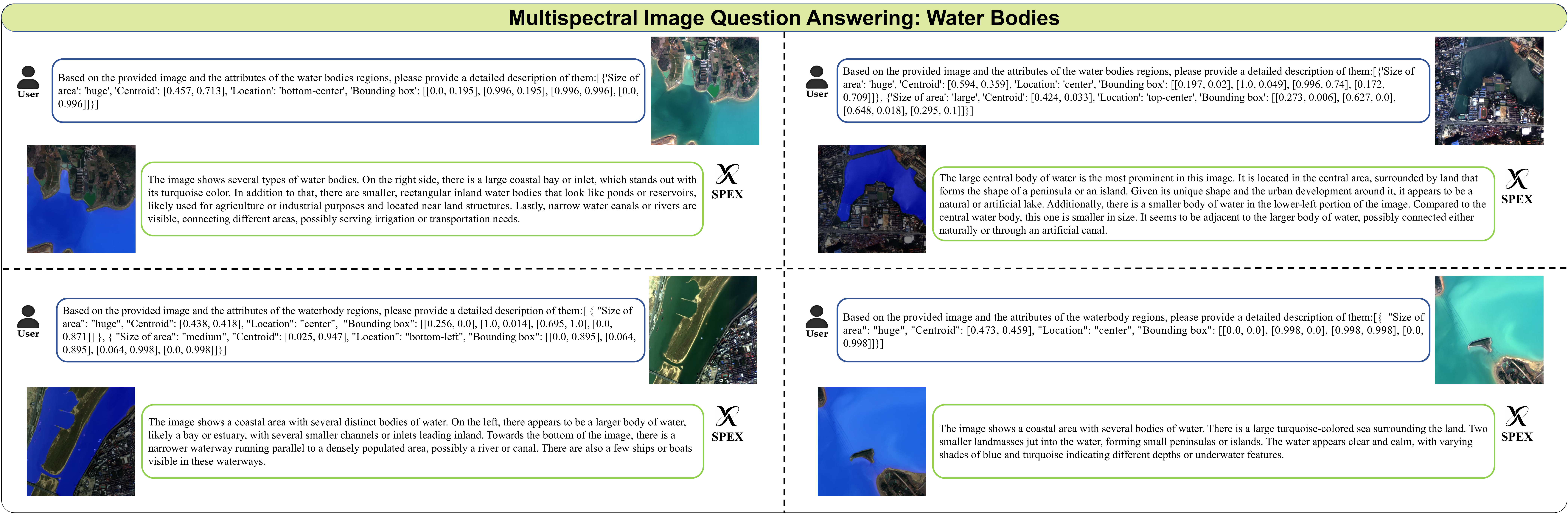} 
    \caption{Examples of water body extraction results, where the obtained masks are highlighted in blue. The samples are from the test set of the GID-15 dataset.}
    \label{fig:example_water}
\end{figure*}%

\begin{figure*}[tbp]
  \centering
    \includegraphics[width=1.0\linewidth]{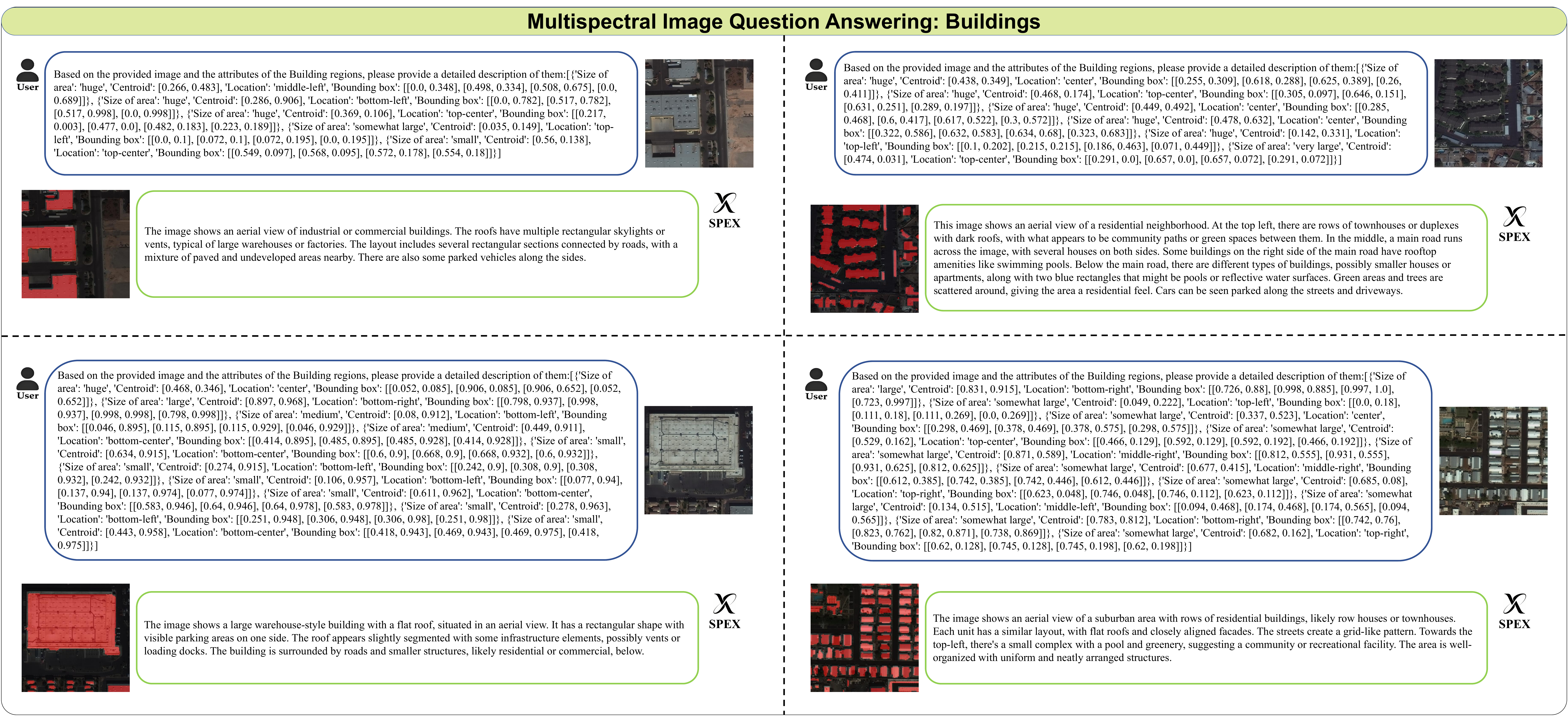} 
    \caption{Examples of building extraction results, where the obtained masks are highlighted in red. The samples are from the test set of the SpaceNet-V2 dataset.}
    \label{fig:example_building}
\end{figure*}%

\begin{figure}[tbp]
  \centering
    \includegraphics[width=1.0\linewidth]{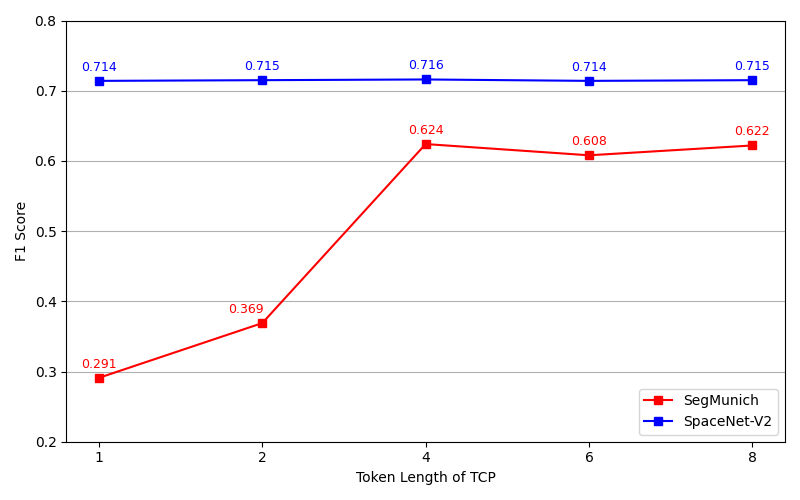} 
    \caption{Performance comparison with different token lengths after
context condensation in the TCP.}
    \label{fig:token_length}
\end{figure}

\begin{table}[t]
\centering
\caption{Ablation study of LLM.}\label{table:w/o-LLM}
    \begin{tabular}{llllll}
    \toprule
    Method & F1 & Recall & Precision & OA & IoU \\
    \midrule
    MVPNet & 0.675 & 0.682 & 0.710 & 0.951  & 0.612 \\
    \midrule
    SPEX & 0.712 & 0.713 & 0.733 & 0.962 & 0.661 \\
    \bottomrule
    \end{tabular}
\end{table}  

\subsubsection{Token Length of TCP}

 We investigate the impact of token length after context condensation in the TCP module, aiming to determine the optimal number of compressed tokens. To this end, we conduct experiments with varying token lengths, as illustrated in Figure~\ref{fig:token_length}. As with other ablation studies, we initially test different token lengths on the SpaceNet-V2 dataset. However, the results show minimal variation. We attribute this to the relatively simple nature of the dataset, as evidenced by the high accuracies achieved by various methods in Table~\ref{table:spacenetv2}, suggesting that even a single token may suffice for effective extraction, and longer tokens provide only marginal gains. To better assess the effect of token length, we further conduct experiments on a more challenging dataset: SegMunich. As shown in Figure~\ref{fig:token_length}, when the token length is set too small (e.g., 1 or 2), SPEX suffers a significant performance drop, likely due to excessive compression that discards essential semantic information. As the token length increases, the performance improves and stabilizes at a length of 4. Considering the trade-off between extraction performance and computational efficiency, we adopt a token length of 4 in the final configuration.

\subsubsection{LLM Scale}

To analyze the impact of LLM scale on SPEX, we conduct ablation studies using Qwen2.5 \cite{yang2024qwen2} models with different parameter sizes, including Qwen2.5-0.5B, Qwen2.5-1.5B, and Qwen2.5-3.0B. The results in Table~\ref{table:LLMType} show that increasing the LLM size from 0.5B to 1.5B leads to consistent improvements in model performance. However, further scaling to 3.0B does not bring additional gains and even causes slight performance degradation on certain metrics, while incurring higher computational cost. This indicates that the semantic guidance required for ground object extraction can be sufficiently captured by a LLM with moderate capacity, and further increasing model size yields diminishing returns. Considering the trade-off between extraction accuracy and computational efficiency, Qwen2.5-1.5B is adopted as the default LLM configuration in SPEX.

\subsubsection{Effect of LLM}

To investigate the impact of the LLM, we further compare the performance of SPEX and MVPNet. Recall that MVPNet is obtained by removing the LLM component and replacing the SAM decoder with a purely visual UperNet decoder, while keeping the hierarchical vision encoder unchanged. Table~\ref{table:w/o-LLM} reports the extraction performance of the two models. It should be noted that, for a comprehensive comparison, each metric in Table~XI is computed as the average of the corresponding results reported in Tables~\ref{table:segmunich}-\ref{table:gid15} (e.g., $0.675 \approx (0.579 + 0.914 + 0.803 + 0.678 + 0.401)/5$). 

As observed, MVPNet consistently exhibits performance degradation across all evaluation metrics. These results indicate that the performance gains of SPEX cannot be solely attributed to hierarchical visual features, but instead stem from the guidance of vision-language representations that incorporate spectral prior information, thereby validating the effectiveness of introducing the LLM component.

\subsection{Visualization}

Finally, we present more examples of generated extraction map–description pairs. The results for vegetation, water bodies, and buildings are shown separately in Figures \ref{fig:example_vegetation}–\ref{fig:example_building}. It can be observed that the generated texts effectively describe the landscape in the multispectral imagery, particularly with regard to the target land cover types, demonstrating the interpretability of our models.

\subsection{Discussion of LLM Hallucination}

Although LLMs may suffer from hallucination in text generation, in SPEX the segmentation is conditioned on the LLM’s hidden representations prior to the text decoding layer, and its decoded textual outputs are not involved in pixel-level prediction or segmentation evaluation. This design largely isolates hallucination effects from the extraction results.

Nevertheless, to improve the reliability and interpretability of textual descriptions for user reference, we introduce a set of inference control mechanisms. Specifically, a decoding strategy with low temperature ($T = 0.3$) and Top-$p$ sampling ($p = 0.9$) is adopted to encourage high-confidence token generation, and a repetition penalty is applied to reduce redundant or unstable outputs. It should be noted that, through token condensation, the proposed TCP module may also help suppress contextual redundancy. Through the coordination of these mechanisms, hallucination effects in textual outputs are largely mitigated, and the risk of unreliable language information propagation is further reduced.

\section{Conclusion}
In this paper, we present SPEX, the first vision-language model specifically designed for instruction-driven, pixel-level land cover extraction in spectral remote sensing imagery. By incorporating spectral priors into the large language model framework, SPEX enables accurate land cover extraction. To support its training, we construct the SPIE dataset, an instruction-following vision-language corpus enriched with structured land cover attributes derived from spectral index computations, as well as descriptive responses generated via auxiliary instruction designs. Extensive experiments across representative land cover categories demonstrate that SPEX outperforms existing methods, achieving state-of-the-art performance while maintaining strong explainability. In conclusion, the SPEX model and the SPIE dataset jointly provide a strong foundation for language-guided interpretation of multispectral remote sensing imagery and hold promise for fostering further research.
\ifCLASSOPTIONcaptionsoff
  \newpage
\fi
\bibliographystyle{IEEEtran}

\normalem
\bibliography{references.bib}

\end{document}